\documentclass{amsart}
\usepackage{enumerate,url,color,booktabs}
\usepackage[pdftex]{graphicx}
\providecommand{\U}[1]{\protect\rule{.1in}{.1in}}
\theoremstyle{plain}
\newtheorem{theorem}{Theorem}[section]

\newtheorem{proposition}[theorem]{Proposition}
\newtheorem{lemma}[theorem]{Lemma}
\newtheorem{definition}[theorem]{Definition}
\begin{document}
\title{Statistical ranking and combinatorial Hodge theory}
\author[X.~Jiang]{Xiaoye~Jiang}
\address{Institute for Computational and Mathematical Engineering, Stanford University,
Stanford, CA 94305}
\email{xiaoyej@stanford.edu}
\author[L.-H.~Lim]{Lek-Heng~Lim}
\address{Department of Mathematics, University of California, Berkeley, CA 94720}
\email{lekheng@berkeley.edu}
\author[Y.~Yao]{Yuan~Yao}
\address{Department of Mathematics, Stanford University, Stanford, CA 94305}
\email{yuany@stanford.edu}
\author[Y.~Ye]{Yinyu~Ye}
\address{Department of Management Science and Engineering, Stanford University,
Stanford, CA 94305}
\email{yinyu-ye@stanford.edu}

\begin{abstract}
We propose a number of techniques for obtaining a global ranking from data
that may be incomplete and imbalanced --- characteristics that are almost
universal to modern datasets coming from e-commerce and internet applications.
We are primarily interested in cardinal data based on scores or ratings though
our methods also give specific insights on ordinal data. From raw ranking
data, we construct pairwise rankings, represented as edge flows on an
appropriate graph. Our statistical ranking method exploits the graph
Helmholtzian, which is the graph theoretic analogue of the Helmholtz operator
or vector Laplacian, in much the same way the graph Laplacian is an analogue
of the Laplace operator or scalar Laplacian. We shall study the graph
Helmholtzian using combinatorial Hodge theory, which provides a way to unravel
ranking information from edge flows. In particular, we show that every edge
flow representing pairwise ranking can be resolved into two orthogonal
components, a gradient flow that represents the $l_{2}$-optimal global ranking
and a divergence-free flow (cyclic) that measures the validity of the global
ranking obtained --- if this is large, then it indicates that the data does
not have a good global ranking. This divergence-free flow can be further
decomposed orthogonally into a curl flow (locally cyclic) and a harmonic flow
(locally acyclic but globally cyclic); these provides information on whether
inconsistency in the ranking data arises locally or globally.

When applied to statistical ranking problems, Hodge decomposition sheds light
on whether a given dataset may be globally ranked in a meaningful way or if
the data is inherently inconsistent and thus could not have any reasonable
global ranking; in the latter case it provides information on the nature of
the inconsistencies. An obvious advantage over the NP-hardness of Kemeny
optimization is that the discrete Hodge decomposition may be easily computed
via a linear least squares regression. We also investigated the $l_{1}$-projection
of edge flows, showing that this has a dual given by correlation
maximization over bounded divergence-free flows, and the $l_{1}$-approximate
sparse cyclic ranking, showing that this has a dual given by correlation
maximization over bounded curl-free flows. We discuss connections with
well-known ordinal ranking techniques such as Kemeny optimization and Borda
count from social choice theory.

\end{abstract}
\maketitle

\section{Introduction\label{intro}}

The problem of ranking in various contexts has become increasingly important
in machine learning. Many datasets require some form of ranking to facilitate
identification of important entries, extraction of principal attributes, and
to perform efficient search and sort operations. Modern internet and
e-commerce applications have spurred an enormous growth in such datasets:
Google's search engine, CiteSeer's citation database, eBay's
feedback-reputation mechanism, Netflix's movie recommendation system, all
accumulate a large volume of data that needs to be ranked.

These modern datasets typically have one or more of the following features
that render traditional ranking methods (such as those in social choice
theory) inapplicable or ineffective: (1) unlike traditional ranking problems
such as votings and tournaments, the data often contains \textit{cardinal
scores} instead of ordinal orderings; (2) the given data is largely
\textit{incomplete} with most entries missing a substantial amount of
information; (3) the data will almost always be \textit{imbalanced} where the
amount of available information varies widely from entry to entry and/or from
criterion to criterion; (4) the given data often lives on a large
\textit{complex network}, either explicitly or implicitly, and the structure
of this underlying network is itself important in the ranking process. These
new features have posed new challenges and call for new techniques. In this
paper we will look at a method that addresses them to some extent.

A fundamental problem here is to globally rank a set of \textit{alternatives}
based on scores given by \textit{voters}. Here the words `alternatives' and
`voters' are used in a generalized sense that depends on the context. For
example, the alternatives may be websites indexed by Google, scholarly
articles indexed by CiteSeer, sellers on eBay, or movies on Netflix; the
voters in the corresponding contexts may be other websites, other scholarly
articles, buyers, or viewers. The `voters' could also refer to groups of
voters: e.g.\ websites, articles, buyers, or viewers grouped respectively by
topics, authorship, buying patterns, or movie tastes. The `voters' could even
refer to something entirely abstract, such as a collection of different
criteria used to judge the alternatives.

The features (1)--(4) can be observed in the aforementioned examples. In the
eBay/Netflix context, a buyer/viewer would assign cardinal scores (1 through 5
stars) to sellers/movies instead of ranking them in an ordinal fashion; the
eBay/Netflix datasets are highly incomplete since most buyers/viewers would
have rated only a very small fraction of the sellers/movies, and also highly
imbalanced since a handful of popular sellers/blockbuster movies will have
received an overwhelming number of ratings while the vast majority will get
only a moderate or small number of ratings. The datasets from Google and
CiteSeer have obvious underlying network structures given by hyperlinks and
citations respectively. Somewhat less obvious are the network structures
underlying the datasets from eBay and Netflix, which come from aggregating the
pairwise comparisons of buyers/movies over all sellers/viewers. Indeed, we
shall see that in all these ranking problems, graph structures naturally arise
from \textit{pairwise comparisons}, irrespective of whether there is an
obvious underlying network (e.g.\ from citation, friendship, or hyperlink
relations) or not, and this serves to place ranking problems of seemingly
different nature on an equal graph-theoretic footing. The incompleteness and
imbalance of the datasets could then be manifested as the (edge) sparsity
structure and (vertex) degree distribution of pairwise comparison graphs.

In collaborative filtering applications, one often encounters a personalized
ranking problem, when one needs to find a global ranking of alternatives that
generates the most consensus within a group of voters who share similar
interests/tastes. While the statistical ranking problem investigated in this
paper plays a fundamental role in such personalized ranking problems, there is
also the equally important problem of clustering voters into interest groups,
which our methods do not address. We would like to stress that in this paper
we only concern ourselves with the ranking problem but not the clustering
problem. So while we have made use of the Netflix prize dataset to motivate
our studies, our paper should not be viewed as an attempt to solve the Netflix
prize problem.

The method that we will use to analyze pairwise rankings, which we represent
as edge flows on a graph, comes from \textit{discrete} or
\textit{combinatorial Hodge theory}. Among other things, combinatorial Hodge
theory provides us with a mean to determine a global ranking that also comes
with a `certificate of reliability' for the validity of this global ranking.
While Hodge theory is well-known to pure mathematicians as a corner stone of
geometry and topology, and to applied mathematician as an important tool in
computational electromagnetics and fluid dynamics, its application to
statistical ranking problems has, to the best of our knowledge, never been
studied\footnote{Nevertheless, Hodge theory has recently found other
applications in statistical learning theory \cite{SmaSma08}.}.

In all our proposed methods, the graph in question has as its vertices the
alternatives to be ranked, voters' preferences are then quantified and
aggregated (we will say how later) into an edge flow on this graph. Hodge
theory then yields an orthogonal decomposition of the edge flow into three
components: a \textit{gradient flow} that is globally acyclic, a
\textit{harmonic flow} that is locally acyclic but globally cyclic, and a
\textit{curl flow} that is locally cyclic. This decomposition is known as the
\textit{Hodge decomposition}. The usefulness of the decomposition lies in the
fact that the gradient flow component induces a global ranking of the
alternatives. Unlike the computationally intractable Kemeny optimal, this may
be easily computed via a linear least squares problem. Furthermore, the
$l_{2}$-norm of the least squares residual, which represents the contribution
from the sum of the remaining curl flow and harmonic flow components,
quantifies the validity of the global ranking induced by the gradient flow
component. If the residual is small, then the gradient flow accounts for most
of the variation in the underlying data and therefore the global ranking
obtained from it is expected to be a majority consensus. On the other hand, if
the residual is large, then the underlying data is plagued with cyclic
inconsistencies (i.e.\ intransitive preference relations of the form $a\succeq
b\succeq c\succeq\dots\succeq z\succeq a$) and one may not assign any
reasonable global ranking to it.

We would like to point out here that cyclic inconsistencies are not
necessarily due to error or noise in the data but may very well be an inherent
characteristic of the data. As the famous impossibility theorems from social
choice theory \cite{Arrow51,Sen98} have shown, inconsistency (or, rather,
intransitivity) is inevitable in any societal preference aggregation that is
sophisticated enough. Social scientists have, through empirical studies,
observed that preference judgement of groups or individuals on a list of
alternatives do in fact exhibit such irrational or inconsistent behavior.
Indeed in any group decision making process, a lack of consensus is the norm
rather than the exception in our everyday experience. This is the well-known
\textit{Condorcet paradox} \cite{Condorcet1785}: the majority prefers $a$ to
$b$ and $b$ to $c$, but may yet prefer $c$ to $a$. Even a single individual
making his own preference judgements could face such dilemma --- if he uses
multiple criteria to rank the alternatives. As such, the cyclic
inconsistencies is intrinsic to any real world ranking data and should be
thoroughly analyzed. Hodge theory again provides a mean to do so. The curl
flow and harmonic flow components of an edge flow quantify respectively the
local and global cyclic inconsistencies.

Loosely speaking, a dominant curl flow component suggests that the
inconsistencies are of a local nature while a dominant harmonic flow component
suggests that they are of a global nature. If most of the inconsistencies come
from the curl (local) component while the harmonic (global) component is
small, then this roughly translates to mean that the ordering of closely
ranked alternatives is unreliable but that of very differently ranked
alternatives is reliable, i.e.\ we cannot say with confidence whether the
ordering of the $27$th, $28$th, $29$th ranked items makes sense but we can say
with confidence that the $4$th, $60$th, $100$th items should be ordered
according to their rank. In other words, Condorcet paradox may well apply to
items ranked closed together but not to items ranked far apart. For example,
if a large number of gourmets (voters) are asked to state their preferences on
an extensive range of food items (alternatives), there may not be a consensus
for their preferences with regard to hamburgers, hot dogs, and pizzas and
there may not be a consensus for their preferences with regard to caviar, foie
gras, and truffles; but there may well be a near universal preference for the
latter group of food items over the former group. In this case, the
inconsistencies will be mostly local and we should expect a large curl flow
component. If in addition the harmonic flow component is small, then most of
the inconsistencies happen locally and we could interpret this to mean that
the global ranking is valid on a coarse scale (ranking different groups of
food) but not on a fine scale (ranking similar food items belonging to a
particular group). We refer the reader to Section~\ref{exp:netflix}
for an explicit example based on the Netflix prize dataset.

When studied in conjunction with robust regression and compressed sensing, the
three orthogonal subspaces given by Hodge decomposition provide other
insights. In this paper we will see two results involving $l_{1}%
$-optimizations where these subspaces provide meaningful and useful
interpretations in the primal-dual way: (a) the $l_{1}$-projection of an edge
flow onto the subspace of gradient flows has a dual problem as the maximal
correlation over bounded cyclic flows, i.e.\ the sum of curl flows and
harmonic flows; (b) the $l_{1}$ -approximation of a sparse cyclic flow, has a
dual problem as the maximal correlation over bounded locally acyclic flows.
These results indicate that the three orthogonal subspaces could arise even in
settings where orthogonality is lost.

\subsection{What's New}

The main contribution of this paper is in the application of Hodge decomposition to the analysis of ranking data. We show that this approach has several attractive features: (i) it generalizes the classical Borda Count method in voting theory to data that may have missing values; (ii) it provides a way to analyze inherent inconsistencies or conflicts in the ranking data; (iii) it is flexible enough to be combined with other techniques: these include other ways to form pairwise rankings reflecting prior knowledge and the use of $l_1$ minimization in place of $l_2$ minimization to encourage robustness or sparsity. Although relatively straightforward and completely natural, the $l_1$ aspects of Hodge theory in Section~\ref{L1} has, to the best of our knowledge, never been discussed before.

We emphasize two conceptual aspects underlying this work that are particularly unconventional: (1) We believe that obtaining a global ranking, which is the main if not the sole objective of all existing work on rank aggregation, gives only an incomplete picture of the ranking data --- one also needs a `certificate of reliability' for the global ranking. Our method provides this certificate by measuring also the local and global inconsistent components of the ranking data. (2) We believe that with the right mathematical model, rank aggregation need not be a computationally intractable task. The model that we proposed in this paper reduces rank aggregation to a linear least squares regression, avoiding usual NP-hard combinatorial optimization problems such as finding Kemeny optima or minimum feedback arc sets.

Hodge and Helmholtz decompositions are of course well-known in mathematics and physics, but usually in a continuous setting where the underlying spaces have the structure of a Riemannian manifold or an algebraic variety. The \textit{combinatorial Hodge theory} that we presented here is arguably a trivial case with the simplest possible underlying space --- a graph. Many of the difficulties in developing Hodge theory in differential and algebraic geometry simply do not surface in our case. However this also makes combinatorial Hodge theory accessible --- the way we developed and presented it essentially requires nothing more than some elementary matrix theory and multivariate calculus. We are unaware of similar treatments in the existing literature and would consider our elementary treatment a minor expository contribution that might help popularize the use of Hodge decomposition and the graph Helmholtzian, possibly to other areas in data analysis and machine learning.

\subsection{Organization of this Paper}

In Section \ref{main} we introduce the main problem and discuss how a pairwise
comparison graph may be constructed from data comprising cardinal scores by
voters on alternatives and how a simple least squares regression may be used
to compute the desired solution. We define the \textit{combinatorial curl}, a
measure of local (triangular) inconsistency for such data, and also the
\textit{combinatorial gradient} and \textit{combinatorial divergence}. Section
\ref{skew} presents a purely matrix-theoretic view of Hodge theory, but at the
expense of some geometric insights. These are covered when we formally
introduce Hodge theory in Section \ref{hodge}. We first remind the reader how
one may construct a $d$-dimensional simplicial complex from any given graph
(the pairwise comparison graph in our case) by simply filling-in all its
$k$-cliques for $k \le d$. Then we will introduce combinatorial Hodge theory
for a general $d$-dimensional simplicial complex but focusing on the $d = 2$
case and its relevance to the ranking problem. In Section \ref{imply} we
discuss the implications of Hodge decomposition applied to ranking, with a
deeper analysis on the least squares method in Section \ref{main}. Section
\ref{L1} extends the analysis to two closely related $l_{1}$-minimization
problems, the $l_{1}$-projection of pairwise ranking onto gradient flows and
the $l_{1}$-approximate sparse cyclic ranking. A discussion of the connections
with Kemeny optimization and Borda count in social choice theory can be found
in Section~\ref{social}. Numerical experiments on three real datasets are
given in Section~\ref{exp} to illustrate some basic ideas in this paper.

\subsection{Notations}

Let $V$ be a finite set. We will adopt the following notation from
combinatorics:
\[
\binom{V}{k} := \text{set of all $k$-element subset of $V$}.
\]
In particular $\binom{V}{2}$ would be the set of all unordered pairs of
elements of $V$ and $\binom{V}{3}$ would be the set of all unordered triples
of elements of $V$ (the sets of ordered pairs and ordered triples will be
denoted $V \times V$ and $V\times V \times V$ as usual). We will not
distinguish between $V$ and $\binom{V}{1}$. Ordered and unordered pairs will
be delimited by parentheses $(i,j)$ and braces $\{i,j\}$ respectively, and
likewise for triples and $n$-tuples in general.

We will use positive integers to label alternatives and voters. Henceforth,
$V$ will always be the set $\{1,\dots,n\}$ and will denote a set of
alternatives to be ranked. In our approach to statistical ranking, these
alternatives would be represented as vertices of a graph. $\Lambda
=\{1,\dots,m\}$ will denote a set of voters. For $i,j\in V$, we write
$i\succeq j$ to mean that alternative $i$ is preferred over alternative $j$.
If we wish to emphasize the preference judgement of a particular voter
$\alpha\in\Lambda$, we will write $i\succeq_{\alpha}j$.

Since our approach mandates that we borrow terminologies from graph theory,
vector calculus, linear algebra, algebraic topology, as well as various
ranking theoretic terms, we think that it would help to summarize some of the
correspondence here.

\begin{center}
{\scriptsize
\begin{tabular}
[c]{l|l|l|l|l}%
\textbf{Graph theory} & \textbf{Linear algebra} & \textbf{Vec.\ calculus} &
\textbf{Topology} & \textbf{Ranking}\\\hline\hline
Function on & Vector in $\mathbb{R}^{n}$ & Potential & $0$-cochain & Score\\
vertices &  & function &  & function\\\hline
Edge flow & Skew-symmetric & Vector field & $1$-cochain & Pairwise\\
& matrix in $\mathbb{R}^{n\times n}$ &  &  & ranking\\\hline
Triangular flow & Skew-symmetric hyper- & Tensor field & $2$-cochain &
Triplewise\\
& -matrix in $\mathbb{R}^{n\times n\times n}$ &  &  & ranking
\end{tabular}
}
\end{center}

As the reader will see, the notions of gradient, divergence, curl, Laplace
operator, and Helmholtz operator from vector calculus and topology will play
important roles in statistical ranking. One novelty of our approach lies in
extending these notions to the other three columns, where most of them have no
well-known equivalent. For example, what we will call a \textit{harmonic
ranking} is central to the question of whether a global ranking is feasible.
This notion is completely natural from the vector calculus or topology
point-of-view, they correspond to solutions of the Helmholtz equation or
homology classes. However, it will be hard to define harmonic ranking directly
in social choice theory without this insight, and we suspect that it is the
reason why the notion of harmonic ranking has never been discussed in existing
studies of ranking in social choice theory and other fields.

\section{Statistical Ranking on Graphs\label{main}}

The main problem discussed in this paper is that of determining a global
ranking from a dataset comprising a set of alternatives ranked by a set of
voters. This is a problem that has received attention in fields including
decision science \cite{Saaty77,Saaty84}, financial economics
\cite{Barber01,Ma06}, machine learning
\cite{KorBell07,CorMohRas07,FreIyeSch98,HerGraObe00}, social choice
\cite{Arrow51,Sen98,Saari00}, statistics
\cite{Dia89,KenGib90,KenSmi40,Mosteller51a,Mosteller51b,Mosteller51c,Noether60}%
, among others. Our objective towards statistical ranking is two-fold: like
everybody else, we want to deduce a global ranking from the data whenever
possible; but in addition to that, we also want to detect when the data does
not permit a statistically meaningful global ranking and in which case
characterize the data in terms of its local and global inconsistencies.

Let $V=\{1,\dots,n\}$ be the set of alternatives to be ranked and $\Lambda=
\{1,\dots, m\}$ be a set of voters. The implicit assumption is that each voter
would have \textit{rated}, i.e.\ assigned cardinal scores or given an ordinal
ordering to, a small fraction of the alternatives. But no matter how
incomplete the rated portion is, one may always convert such ratings into
\textit{pairwise rankings} that has no missing values as follows. For each
voter $\alpha\in\Lambda$, the \textit{pairwise ranking matrix} of $\alpha$ is
a skew-symmetric matrix $Y^{\alpha}\in\mathbb{R}^{n \times n}$, i.e.\ for each
ordered pair $(i,j)\in V \times V$, we have
\[
Y^{\alpha}_{ij} = -Y^{\alpha}_{ji}.
\]
Informally, $Y^{\alpha}_{ij}$ measures the `degree of preference' of the $i$th
alternative over the $j$th alternative held by the $\alpha$th voter. Studies
of ranking problems in different disciplines have led to rather different ways
of quantifying such `degree of preference'. In Section \ref{exmp:netflix}, we
will see several ways of defining $Y^{\alpha}_{ij}$ (as score difference,
score ratio, and score ordering) coming from decision science, machine
learning, social choice theory, and statistics. If the voter $\alpha$ did not
compare alternatives $i$ and $j$, then $Y^{\alpha}_{ij}$ is considered a
missing value and set to be $0$ for convenience; this manner of handling
missing values allows $Y^{\alpha}$ to be a skew-symmetric matrix for each
$\alpha\in\Lambda$. Nevertheless we could have assigned any arbitrary value or
a non-numerical symbol to represent missing values, and this would have not
affected our algorithmic results because of our use of the following weight function.

Define the \textit{weight function} $w:\Lambda\times V\times V\rightarrow
\lbrack0,\infty)$ as the indicator function
\[
w_{ij}^{\alpha}=w(\alpha,i,j)=%
\begin{cases}
1 & \text{if $\alpha$ made a pairwise comparison for $\{i,j\}$},\\
0 & \text{otherwise}.
\end{cases}
\]
Therefore $w_{ij}^{\alpha}=0$ iff $Y_{ij}^{\alpha}$ is a missing value. Note
that $W^{\alpha}=[w_{ij}^{\alpha}]$ is a symmetric $\{0,1\}$-valued matrix;
but more generally, $w_{ij}^{\alpha}$ may be chosen as the capacity (in the
graph theoretic sense) if there are multiple comparisons of $i$ and $j$ by
voter $\alpha$. The pairs $(i,j)$ for which $w(\alpha,i,j)=1$ for some
$\alpha\in\Lambda$ are known as \textit{crucial pairs} in the machine learning
literature (we thank the reviewers for pointing this out).

Our general paradigm for statistical ranking is to minimize a weighted sum of
pairwise loss of a global ranking on the given data over a model class
$\mathcal{M}$ of all global rankings. We begin with a simple sum-of-squares
loss function,
\begin{equation}
\min_{X\in\mathcal{M}_{G}}\sum\nolimits_{\alpha,i,j}w_{ij}^{\alpha}%
(X_{ij}-Y_{ij}^{\alpha})^{2}, \label{eq:main1}%
\end{equation}
where the \textit{model class} $\mathcal{M}_{G}$ is a subset of the
skew-symmetric matrices,
\begin{equation}
\mathcal{M}_{G}=\{X\in\mathbb{R}^{n\times n}\mid X_{ij}=s_{j}-s_{i}%
,s:V\rightarrow\mathbb{R}\}. \label{eq:M}%
\end{equation}
Any $X\in\mathcal{M}_{G}$ induces a global ranking on the alternatives
$1,\dots,n$ via the rule $i\succeq j$ iff $s_{i}\geq s_{j}$. Note that ties,
i.e.\ $i\succeq j$ and $j\succeq i$, are allowed and this happens precisely
when $s_{i}=s_{j}$.

For ranking data given in terms of cardinal scores, this simple scheme
preserves the magnitudes of the ratings, instead of merely the ordering, when
we have globally consistent data (see Definition \ref{def:consistency}).
Moreover, it may also be computed more easily than many other loss functions
(though the computational cost depends also on the choice of $\mathcal{M}$).
This simple scheme is not as restrictive as it first seems. For example,
Kemeny optimization in classical social choice theory may be realized as a
special case where $Y^{\alpha}_{ij} \in\{\pm1\}$ and $\mathcal{M}$ is the
\textit{Kemeny} model class,
\begin{equation}
\label{eq:Mk}\mathcal{M}_{K} := \{ X \in\mathbb{R}^{n \times n} \mid X_{ij} =
\operatorname*{sign} (s_{j} - s_{i}), s: V\to\mathbb{R}\}.
\end{equation}
The function $\operatorname*{sign}:\mathbb{R}\to\{\pm1\}$ takes nonnegative
numbers to $1$ and negative numbers to $-1$. A binary valued $Y^{\alpha}_{ij}
$ is the standard scenario in binary pairwise comparisons
\cite{ailon05,Arrow51,David69,HerGraObe00,KenSmi40}; in this context, a global
ranking is usually taken to be synonymous as a Kemeny optimal. We will discuss
Kemeny optimization in greater details in Section \ref{social}.

\subsection{Pairwise Comparison Graphs and Pairwise Ranking
Flows\label{sec:Pairwise}}

A graph structure arises naturally from ranking data as follows. Let $G=(V,E)$
be an undirected graph whose vertex set is $V$, the set of alternatives to be
ranked, and whose edge set is
\begin{equation}
\label{eq:Edge}E=\bigl\{\{i,j\} \in\tbinom{V}{2} \bigm| \textstyle{\sum
_{\alpha}}w^{\alpha}_{ij} >0 \bigr\},
\end{equation}
i.e.\ the set of pairs $\{i,j\}$ where pairwise comparisons have been made. We
call such $G$ a \textit{pairwise comparison graph}. One can further associate
weights on the edges as capacity, e.g.\ $w_{ij} = \sum_{\alpha}w^{\alpha}%
_{ij}$.

A pairwise ranking can be viewed as \textit{edge flows} on $G$, i.e.\ a
function $X:V\times V\rightarrow\mathbb{R}$ that satisfies
\begin{align}
X(i,j)  &  =-X(j,i) &  &  \text{if }\{i,j\} \in E,\nonumber\\
X(i,j)  &  =0 &  &  \text{otherwise.} \label{eq:EdgeFlows}%
\end{align}
It is clear that a skew-symmetric matrix $[X_{ij}]$ induces an edge flow and
vice versa. So henceforth we will not distinguish between edge flows and
skew-symmetric matrices and will often write $X_{ij}$ in place of $X(i,j)$.

We will now borrow some terminologies from vector calculus. An edge flow of
the form $X_{ij}=s_{j}-s_{i}$, i.e.\ $X\in\mathcal{M}_{G}$, can be regarded as
the \textit{gradient} of a function $s:V\rightarrow\mathbb{R}$, which will be
called a \textit{potential} function (or \textit{negative potential},
depending on sign convention). In the context of ranking, a potential function
is a \textit{score function} or \textit{utility function} on the set of
alternatives, assigning a score $s(i)=s_{i}$ to alternative $i$. Note that any
such function defines a \textit{global ranking} as discussed after
\eqref{eq:M}. To be precise, we define gradient as follows.

\begin{definition}
\label{def:grad} The \textbf{combinatorial gradient} operator maps a potential
function on the vertices $s:V\rightarrow\mathbb{R}$ to an edge flow
$\operatorname*{grad}s:V\times V\rightarrow\mathbb{R}$ via
\begin{equation}
(\operatorname*{grad}s)(i,j)=s_{j}-s_{i}. \label{eq:grad}%
\end{equation}
An edge flow that has this form will be called a \textbf{gradient flow}.
\end{definition}

In other words, the combinatorial gradient takes global rankings to pairwise
rankings. Pairwise rankings that arise in this manner will be called
\textit{globally consistent} (formally defined in Definition
\ref{def:consistency}). Given a globally consistent pairwise ranking $X$, we
can easily solve $\operatorname*{grad}(s) = X$ to determine a score function
$s$ (up to an additive constant), and from $s$ we can obtain a global ranking
of the alternatives in the manner described after \eqref{eq:M}. Observe that
the set of all globally consistent pairwise rankings in \eqref{eq:M} may be
written as $\mathcal{M}_{G} =\{\operatorname*{grad} s \mid s:V\to\mathbb{R}\}
= \operatorname*{im}(\operatorname*{grad})$.

For convenience, we will drop the adjective `combinatorial' from
`combinatorial gradient'. We may sometimes also drop the adjective `pairwise'
in `globally consistent pairwise ranking' when there is no risk of confusion.

The optimization problem \eqref{eq:main1} can be rewritten in the form of a
weighted $l_{2}$-minimization on a pairwise comparison graph
\begin{equation}
\label{eq:main2}\min_{X\in\mathcal{M}_{G}} \| X - \bar{Y} \|_{2,w}^{2} =
\min_{X\in\mathcal{M}_{G}} \Bigl[ \sum\nolimits_{\{i,j\}\in E} w_{ij} (X_{ij}
- \bar{Y}_{ij} )^{2}\Bigr]
\end{equation}
where
\begin{equation}
\label{eq:W}w_{ij} := \textstyle{\sum_{\alpha}} w^{\alpha}_{ij} \quad
\text{and}\quad\bar{Y}_{ij} := \frac{\sum_{\alpha}w^{\alpha}_{ij} Y^{\alpha
}_{ij} } {\sum_{\alpha}w^{\alpha}_{ij}}.
\end{equation}
An optimizer thus corresponds to an $l_{2}$-projection of a pairwise ranking
edge flow $\bar{Y}$ onto the space of gradient flows. We note that $W = [w_{ij}]
= \sum_{\alpha}W^{\alpha}$ is a symmetric nonnegative-valued matrix.
This choice of $W$ is not intended to be rigid. One could for example define
$W$ to incorporate prior knowledge of the relative importance of the paired comparisons
as judged by the voters.

An interesting variation of this scheme is an analogous $l_{1}$-projection
onto the space of gradient flows,
\begin{equation}
\min_{X\in\mathcal{M}_{G}}\Vert X-\bar{Y}\Vert_{1,w}=\min_{X\in\mathcal{M}%
_{G}}\Bigl[\sum\nolimits_{\{i,j\}\in E}w_{ij}\lvert X_{ij}-\bar{Y}_{ij}%
\rvert\Bigr]. \label{eq:mainl1}%
\end{equation}
Its solutions are more robust to outliers or large deviations in $\bar{Y}%
_{ij}$ as \eqref{eq:mainl1} may be regarded as the \textit{least absolute
deviation} (LAD) method in robust regression. We will discuss this problem in
greater details in Section \ref{robust}.

Combinatorial Hodge theory will provide a geometric interpretation of the
optimizer and residuals of \eqref{eq:main2} as well as further insights on
\eqref{eq:mainl1}. Before going deeper into the analysis of such optimization
problems, we present several examples of pairwise ranking arising from applications.

\subsection{Pairwise Rankings\label{exmp:pairwise}}

Humans are unable to make accurate preference judgement on even moderately
large sets. In fact, it has been argued that most people can rank only between
5 to 9 alternatives at a time \cite{Swt}. This is probably why many rating
scales (e.g.\ the ones used by Amazon, eBay, Netflix, YouTube) are all based
on a $5$-star scale. Hence one expects large human-generated ranking data to
be at best partially ordered (with chains of lengths about $5$ to $9$, if
\cite{Swt} is accurate). For most people, it is a harder task to rank or rate
$20$ movies than to compare the movies a pair at a time. In certain settings
such as tennis tournaments and wine tasting, only pairwise comparisons are
possible. Pairwise comparison methods, which involve the smallest partial
rankings, is thus natural for analyzing ranking data.

Pairwise comparisons also help reduce bias due to the arbitrariness of rating
scale by adopting a relative measure. As we will see in Section
\ref{exmp:netflix}, pairwise comparisons provide a way to handle missing
values, which are expected because of the general lack of incentives or
patience for a human to process a large dataset. For these reasons, pairwise
comparison methods have been popular in psychology, statistics, and social
choice theory \cite{Thurstone27,KenSmi40,David69,Saaty77,Arrow51}. Such
methods are also getting increasing attention from the machine learning
community as they may be adapted for studying classification problems
\cite{HasTib98,FreIyeSch98,HerGraObe00}. We will present two very different
instances where pairwise rankings arise: recommendation systems and exchange
economic systems.

\subsubsection{Recommendation systems\label{exmp:netflix}}

The generic scenario in recommendation systems is that there are $m$ voters
rating $n$ alternatives. For example, in the Netflix context, viewers will
rate a movie on a scale of 5 stars \cite{KorBell07}; in financial markets,
analysts will rate a stock or a security by 5 classes of recommendations
\cite{Barber01}. In these cases, we let $A=[a_{\alpha i}]\in\mathbb{R}%
^{m\times n}$ represent the voter-alternative matrix. $A$ typically has a
large number of missing values; for example, the dataset that Netflix released
for its prize competition contains a viewer-movie matrix with 99\% of its
values missing. The standard problem here is to predict these missing values
from the given data but we caution the reader again that this is \textit{not}
the problem addressed in our paper. Instead of estimating the missing values
of $A$, we want to learn a global ranking of the alternatives from $A$,
without having to first estimate the missing values.

Even though the matrix $A$ may be highly incomplete, we may aggregate over all
voters to get a pairwise ranking matrix using one of the four following methods.

\begin{enumerate}
\item \textbf{Arithmetic mean of score differences:} The score difference
refers to $Y^{\alpha}_{ij} = a_{\alpha j}-a_{\alpha i}$. The arithmetic mean
over all customers who have rated both $i$ and $j$ is
\[
\bar{Y}_{ij}=\frac{\sum_{\alpha}(a_{\alpha j}-a_{\alpha i})}{\#\{\alpha\mid
a_{\alpha i},a_{\alpha j}\text{ exist}\}}.
\]
This is translation invariant.

\item \textbf{Geometric mean of score ratios:} Assuming $A>0$. The score ratio
refers to $Y^{\alpha}_{ij} = a_{\alpha j}/a_{\alpha i}$. The (log) geometric
mean over all customers who have rated both $i$ and $j$ is
\[
\bar{Y}_{ij}=\frac{\sum_{\alpha}(\log a_{\alpha j} -\log a_{\alpha i}
)}{\#\{\alpha\mid a_{\alpha i},a_{\alpha j}\text{ exist}\}}.
\]
This is scale invariant.

\item \textbf{Binary comparison:} Here $Y^{\alpha}_{ij} = \operatorname*{sign}%
(a_{\alpha j} - a_{\alpha i} )$. Its average is the probability difference
that the alternative $j$ is preferred to $i$ than the other way round,
\[
\bar{Y}_{ij}=\Pr\{\alpha\mid a_{\alpha j}>a_{\alpha i}\}-\Pr\{\alpha\mid
a_{\alpha j}<a_{\alpha i}\}.
\]
This is invariant up to a monotone transformation.

\item \textbf{Logarithmic odds ratio:} As in the case of binary comparison,
except that we adopt a logarithmic scale
\[
\bar{Y}_{ij}=\log\frac{\Pr\{\alpha\mid a_{\alpha j}\geq a_{\alpha i}\}}%
{\Pr\{\alpha\mid a_{\alpha j}\leq a_{\alpha i}\}}.
\]
This is also invariant up to a monotone transformation.
\end{enumerate}

Each of these four statistics is a form of ``average pairwise ranking'' over
all voters. The first model leads to the concept of \textit{position-rules} in
social choice theory \cite{Saari00} and it has also been used in machine
learning recently \cite{CorMohRas07}. The second model has appeared in
multi-criteria decision theory \cite{Saaty77}. The third and fourth models are
known as \textit{linear model} \cite{Noether60} and \textit{Bradley-Terry
model} \cite{BraTer52} respectively in the statistics and psychology
literature. There are other plausible choices for defining $\bar{Y}_{ij}$,
e.g.\ \cite{Thurstone27,Mosteller51a,Mosteller51b,Mosteller51c}, but we will
not discuss more of them here. It suffices to note that there is a rich
variety of techniques to preprocess raw ranking data into the pairwise ranking
edge flow $\bar{Y}_{ij}$ that serves as input to our Hodge theoretic method.
However, it should be noted that the $l_{2}$- and $l_{1}$-optimization on
graphs in \eqref{eq:main2} and \eqref{eq:mainl1} may be applied with any of
the four choices above since only the knowledge of $\bar{Y}_{ij}$ is required
but the sum-of-squares and Kemeny optimization in \eqref{eq:main1} and
\eqref{eq:Mk} require the original score difference or score order data be
known for each voter.

\subsubsection{Exchange economic systems\label{exmp:currency}}

A purely exchange economic system may be described by a graph $G=(V,E)$ with
vertex set $V=\{1,\dots,n\}$ representing the $n$ goods and edge set $E
\subseteq\binom{V}{2}$ representing feasible pairwise transactions. If the
market is complete in the sense that every pair of goods is exchangeable, then
$G$ is a complete graph. Suppose the exchange rate between the $i$th and $j$th
goods is given by
\[
1\text{ unit }i=a_{ij}\text{ unit }j,\quad a_{ij}>0.
\]
Then the exchange rate matrix $A=[a_{ij}]$ is a \textit{reciprocal matrix}
(possibly with missing values), i.e.\ $a_{ij}=1/a_{ji}$ for all $i,j \in V$.
The reciprocal matrix was first used in the studies of paired preference
aggregation by Saaty \cite{Saaty77}; it was also used by Ma \cite{Ma06} to
study currency exchange markets. A pricing problem here is to look for a
\textit{universal equivalent} which measures the values of goods (this is in
fact an abstraction of the concept of money), i.e.\ $\pi:V\rightarrow
\mathbb{R}$ such that
\[
a_{ij}=\frac{\pi_{j}}{\pi_{i}}.
\]
In complete markets where $G$ is a complete graph, there exists a universal
equivalent if and only if the market is \textit{triangular arbitrage-free},
i.e.\ $a_{ij}a_{jk}=a_{ik}$ for all distinct $i,j,k\in V$; since in this case
the transaction path $i\rightarrow j\rightarrow k$ provides no gain nor loss
over a direct exchange $i\rightarrow k$.

Such purely exchange economic system is equivalent to pairwise ranking via the
logarithmic map,
\[
X_{ij}=\log a_{ij}.
\]
The triangular arbitrage-free condition is then equivalent to the transitivity
condition in \eqref{eq:transitive}, i.e.
\[
X_{ij}+X_{jk}+X_{ki}=0.
\]
So asking if a universal equivalent exists is the same as asking if a global
ranking $s:V\rightarrow\mathbb{R}$ exists so that $X_{ij}=s_{j}-s_{i}$ with
$s_{i}=\log\pi_{i}$.

\subsection{Measure of Triangular Inconsistency: combinatorial curl}

Upon constructing pairwise rankings from the raw data, we need a statistics to
quantify the inconsistency in the pairwise rankings. Again we will borrow a
terminology from vector calculus and define a notion of \textit{combinatorial
curl} as a measure of triangular inconsistency.

Given a pairwise ranking represented as an edge flow $X$ on a graph $G=(V,E)$,
we expect the following `consistency' property: following a loop $i\rightarrow
j\rightarrow\dots\rightarrow i$ where each edge is in $E$, the amount of the
scores raised should be equal to the amount of the scores lowered; so after a
loop of comparisons we should return to the same score on the same
alternative. Since the simplest loop is a triangular loop $i\rightarrow
j\rightarrow k\rightarrow i$, the `basic unit' of inconsistency should be
triangular in nature and this leads us to the combinatorial curl in Definition
\ref{def:curl}.

We will first define a notion analogous to edge flows. The \textit{triangular
flow} on $G$ is a function $\Phi: V \times V \times V \to\mathbb{R}$ that
satisfies
\[
\Phi(i,j,k) = \Phi(j,k,i) = \Phi(k,i,j) = -\Phi(j,i,k) = - \Phi(i,k,j) = -
\Phi(k,j,i),
\]
i.e.\ an odd permutation of the arguments of $\Phi$ changes its sign while an
even permutation preserves its sign\footnote{A triangular flow is an
alternating $3$-tensor and may be represented as a skew-symmetric hypermatrix
$[\Phi_{ijk}] \in\mathbb{R}^{n\times n \times n}$, much like an edge flow is
an alternating $2$-tensor and may be represented by a skew-symmetric matrix
$[X_{ij}] \in\mathbb{R}^{n \times n}$. We will often write $\Phi_{ijk}$ in
place of $\Phi(i,j,k)$.}. A triangular flow describes \textit{triplewise
rankings} in the same way an edge flow describes pairwise rankings.

\begin{definition}
\label{def:curl} Let $X$ be an edge flow on a graph $G=(V,E)$. Let
\[
T(E) :=\bigl\{\{i,j,k\} \in\tbinom{V}{n} \bigm| \{i,j\},\{j,k\},\{k,i\}\in E
\bigr\}
\]
be the collection of triangles with every edge in $E$. We define the
\textbf{combinatorial curl} operator that maps edge flows to triangular flows
by
\begin{equation}
\label{eq:curl}(\operatorname*{curl} X)(i,j,k)=
\begin{cases}
X_{ij}+X_{jk}+X_{ki} & \text{if }\{i,j,k\}\in T(E) ,\\
0 & \text{otherwise.}%
\end{cases}
\end{equation}

\end{definition}

In other words, the combinatorial curl takes pairwise rankings to triplewise
rankings. Again, we will drop the adjective `combinatorial' when there is no
risk of confusion. The skew-symmetry of $X$, i.e.\ $X_{ij}=-X_{ji}$,
guarantees that $\operatorname*{curl}X$ is a triangular flow, i.e.
\begin{gather*}
(\operatorname*{curl}X)(i,j,k)=(\operatorname*{curl}%
X)(j,k,i)=(\operatorname*{curl}X)(k,i,j)\\
=-(\operatorname*{curl}X)(j,i,k)=-(\operatorname*{curl}%
X)(i,k,j)=-(\operatorname*{curl}X)(k,j,i).
\end{gather*}

The curl of a pairwise ranking measures its triangular inconsistency. This
extends the \textit{consistency index} of Kendall and Smith \cite{KenSmi40},
which counts the number of circular triads, from ordinal settings to cardinal
settings. Note that for binary pairwise ranking where $X_{ij} \in\{\pm1\}$,
the absolute value $|(\operatorname*{curl} X)(i,j,k)|$ may only take two
values, $1$ or $3$. The triangle $\{i,j,k \} \in T(E)$ contains a cyclic
ranking or \textit{circular triad} if and only if $|(\operatorname*{curl}
X)(i,j,k)| = 3$. If $G$ is a complete graph, the number of circular triads has
been shown \cite{KenSmi40} to be
\[
N = \frac{n}{24} (n^{2} - 1) - \frac{1}{8} \sum\nolimits_{i}\left[
\sum\nolimits_{j} X_{ij} \right]  ^{2}.
\]

For ranking data given in terms of cardinal scores and that is generally
incomplete, curl plays an extended role in addition to just quantifying the
triangular inconsistency. We now formally define some ranking theoretic
notions in terms of the combinatorial gradient and combinatorial curl.

\begin{definition}
\label{def:consistency} Let $X : V\times V \to\mathbb{R}$ be a pairwise
ranking edge flow on a pairwise comparison graph $G = (V,E)$.

\begin{enumerate}
\item $X$ is called \textbf{consistent} on $\{ i,j,k \} \in T(E)$ if it is
curl-free on $\{ i,j,k \}$, i.e.
\[
(\operatorname*{curl} X)(i,j,k) = X_{ij}+X_{jk}+X_{ki}=0.
\]
Note that this implies that $\operatorname*{curl}(X)(\sigma(i),\sigma
(j),\sigma(k)) =0$ for every permutation $\sigma$.

\item $X$ is called \textbf{globally consistent} if it is a gradient flow of a
score function, i.e.
\[
X = \operatorname*{grad} s \quad\text{for some } s : V \to\mathbb{R}.
\]

\item $X$ is called \textbf{locally consistent} or \textbf{triangularly
consistent} if it is curl-free on every triangle in $T(E)$, i.e.\ every
$3$-clique of $G$.
\end{enumerate}
\end{definition}

Clearly any gradient flow must be curl-free everywhere, i.e.\ the well-known
identity in vector calculus
\[
\operatorname*{curl} \circ\operatorname*{grad} = 0
\]
is also true for combinatorial curl and combinatorial gradient (a special case
of Lemma \ref{lem:close}). So global consistency implies local consistency. A
qualified converse may be deduced from the Hodge decomposition theorem (see
also Theorem \ref{thm:consistency}): a curl-free flow on a complete graph must
necessarily be a gradient flow, or putting it another way, a locally
consistent pairwise ranking must necessarily be a globally consistent pairwise
ranking when there are no missing values, i.e.\ if the pairwise comparison
graph is a complete graph (every pair of alternatives has been compared).

\begin{figure}[ptb]
\centering
\resizebox{4cm}{!}{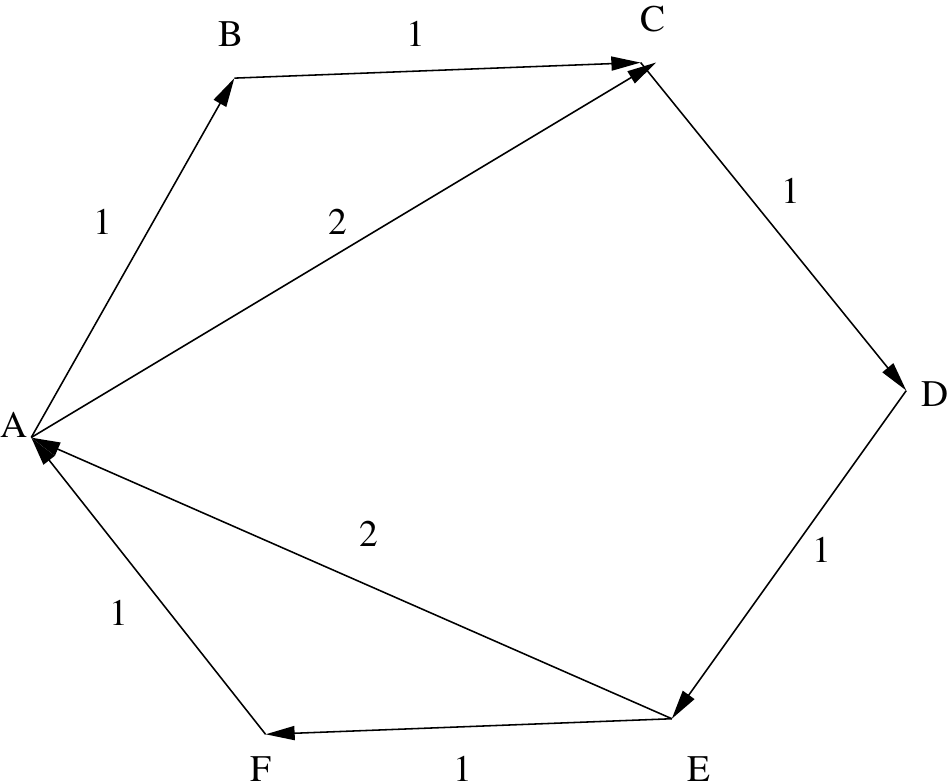}\caption{A harmonic
pairwise ranking, which is locally consistent on every triangles but
inconsistent along the loop $A\to B\to C\to D\to E\to F\to A$.}%
\label{fig:harmonic}%
\end{figure}

When $G$ is an incomplete graph, the condition that $X$ is curl-free on every
triangle in the graph will not be enough to guarantee that it is a gradient
flow. The reason lies in that curl only takes into account the triangular
inconsistency; but since there are missing edges in the pairwise comparison
graph $G$, it is possible that non-triangular cyclic rankings of lengths
greater than three can occur. For example, Figure \ref{fig:harmonic} shows a
pairwise ranking that is locally consistent on every triangle but globally
inconsistent, since it contains a cyclic ranking of length six. Fortunately,
Hodge decomposition theorem will tell us that all such cyclic rankings lie in
a subspace of \textit{harmonic rankings}, which can be characterized as the
kernel of some combinatorial Laplacians. 

\section{A Matrix Theoretic View of Hodge Decomposition\label{skew}}

We will see in this section that edge flows, gradient flows, harmonic flows,
and curl flows can all be represented as specially structured skew-symmetric
matrices. In this framework, the Hodge decomposition theorem may be viewed as
an orthogonal direct sum decomposition of the space of skew-symmetric matrices
into three subspaces. A formal treatment of combinatorial Hodge theory will be
given in Section \ref{hodge}.

Recall that a matrix $X \in\mathbb{R}^{n \times n}$ is said to be
\textit{skew-symmetric} if $X_{ij} = -X_{ji}$ for all $i,j \in V := \{
1,\dots,n \}$. One knows from linear algebra that any square matrix $A$ may be
written uniquely as a sum of a symmetric and a skew-symmetric matrix,
\[
A = \tfrac{1}{2}(A + A^{\top}) + \tfrac{1}{2}(A - A^{\top}).
\]
We will denote\footnote{More common notations for $\mathcal{A}$ are
$\mathfrak{so}_{n}(\mathbb{R})$ (Lie algebra of $\operatorname*{SO}(n)$) and
$\wedge^{2}(\mathbb{R}^{n})$ (second exterior product of $\mathbb{R}^{n}$) but
we avoided these since we use almost no Lie theory and exterior algebra.}
\[
\mathcal{A} :=\{ X \in\mathbb{R}^{n \times n}\mid X^{\top}= -X \},
\quad\text{and} \quad\mathcal{S} := \{ X \in\mathbb{R}^{n \times n}\mid
X^{\top}= X \}.
\]
It is perhaps interesting to note that semidefinite programming takes place in
the cone of symmetric positive definite matrices in $\mathcal{S}$ but the
optimization problems in this paper take place in the exterior space
$\mathcal{A}$.

One simple way to construct a skew-symmetric matrix is to take a vector $s =
[s_{1},\dots,s_{n}]^{\top}\in\mathbb{R}^{n}$ and define $X$ by
\[
X_{ij} := s_{i} - s_{j}.
\]
Note that if $X \ne0$, then $\operatorname*{rank}(X) = 2$ since it can be
expressed as $se^{\top}- es^{\top}$ with $e := [1,\dots,1]^{\top}\in
\mathbb{R}^{n}$. These are in a sense the simplest type of skew-symmetric
matrices --- they have the lowest rank possible for a non-zero skew-symmetric
matrix (recall that the rank of a skew-symmetric matrix is necessarily even).
In this paper, we will call these \textit{gradient} matrices and denote them
collectively by $\mathcal{M}_{G}$,
\[
\mathcal{M}_{G} := \{ X \in\mathcal{A} \mid X_{ij} = s_{i} - s_{j} \text{ for
some } s \in\mathbb{R}^{n} \}.
\]

For $T \subseteq\binom{V}{3}$, we define the set of $T$-\textit{consistent}
matrices as
\begin{equation}
\label{eq:transitive}\mathcal{M}_{T} := \{ X \in\mathcal{A} \mid X_{ij} +
X_{jk} + X_{ki} = 0 \text{ for all } \{i,j,k\} \in T \}.
\end{equation}
We can immediately observe every $X \in\mathcal{M}_{G}$ is $T$%
-\textit{consistent} for any $T \subseteq\binom{V}{3}$, i.e.\ $\mathcal{M}_{G}
\subseteq\mathcal{M}_{T}$. Conversely, a matrix $X$ that satisfies
\[
X_{ij} + X_{jk} + X_{ki} = 0 \quad\text{for every triple } \{i,j,k\} \in
\binom{V}{3}.
\]
is necessarily a gradient matrix, i.e.\
\begin{equation}
\label{eq:gradtrans}\mathcal{M}_{G} = \mathcal{M}_{ \binom{V}{3}}.
\end{equation}

Given $T\subseteq\binom{V}{3}$, it is straightforward to verify that both
$\mathcal{M}_{G}$ and $\mathcal{M}_{T}$ are subspaces of $\mathbb{R}^{n\times
n}$. The preceding discussions then imply the following subspace relations:
\begin{equation}
\mathcal{M}_{G}\subseteq\mathcal{M}_{T}\subseteq\mathcal{A}. \label{eq:subsp}%
\end{equation}
Since these are strict inclusions in general, several complementary subspaces
arise naturally. With respect to the usual inner product $\langle
X,Y\rangle=\operatorname*{tr}(X^{\top}Y)=\sum_{i,j}X_{ij}Y_{ij}$, we obtain
orthogonal complements of $\mathcal{M}_{G}$ and $\mathcal{M}_{T}$ in
$\mathcal{A}$ as well as the orthogonal complement of $\mathcal{M}_{G}$ in
$\mathcal{M}_{T}$, which we denote by $\mathcal{M}_{H}$:
\[
\mathcal{A}=\mathcal{M}_{G}\oplus\mathcal{M}_{G}^{\perp},\qquad\mathcal{A}%
=\mathcal{M}_{T}\oplus\mathcal{M}_{T}^{\perp},\qquad\mathcal{M}_{T}%
=\mathcal{M}_{G}\oplus\mathcal{M}_{H}.
\]
We will call the elements of $\mathcal{M}_{H}$ \textit{harmonic} matrices as
we shall see that they are discrete analogues of solutions to the Laplace
equation (or, more accurately, the Helmholtz equation). An alternative
characterization of $\mathcal{M}_{H}$ is
\[
\mathcal{M}_{H}=\mathcal{M}_{T}\cap\mathcal{M}_{G}^{\perp},
\]
which may be viewed as a discrete analogue of the condition of being
simultaneously curl-free and divergence-free. More generally, this discussion
applies to any weighted inner product $\langle X,Y\rangle_{w}=\sum_{i,j}%
w_{ij}X_{ij}Y_{ij}$. The five subspaces $\mathcal{M}_{G},\mathcal{M}%
_{T},\mathcal{M}_{H},\mathcal{M}_{T}^{\perp},\mathcal{M}_{G}^{\perp}$ of
$\mathcal{A}$ play a central role in our techniques. As we shall see later,
the Helmholtz decomposition in Theorem \ref{thm:helmholz} may be viewed as the
orthogonal direct sum decomposition
\[
\mathcal{A}=\mathcal{M}_{G}\oplus\mathcal{M}_{H}\oplus\mathcal{M}_{T}^{\perp
}.
\]

\section{Combinatorial Hodge Theory\label{hodge}}

In this section we will give a brief introduction to combinatorial Hodge
theory, paying special attention to its relevance in statistical ranking. One
may wonder why we do not rely on our relatively simple matrix view in Section
\ref{skew}. The reasons are two fold: firstly, important geometric insights
are lost when the actual motivations behind the matrix picture are
disregarded; and secondly, the matrix approach applies only to the case of
$2$-dimensional simplicial complex but combinatorial Hodge theory extends to
any $k$-dimensional simplicial complex. While so far we did not use any
simplicial complex of dimension higher than $2$ in our study of statistical
ranking, it is conceivable that higher-dimensional simplicial complex could
play a role in future studies.

\subsection{Extension of Pairwise Comparison Graph to Simplicial Complex}

Let $G=(V,E)$ be a pairwise comparison graph. To characterize the triangular
inconsistency or curl, one needs to study the triangles formed by the
$3$-cliques\footnote{Recall that a $k$-clique of $G$ is just a complete
subgraph of $G$ with $k$ vertices.}, i.e.\ the set
\[
T(E) := \bigl\{\{i,j,k\} \in\tbinom{V}{3} \bigm| \{i,j\},\{j,k\},\{k,i\}\in E
\bigr\}.
\]
A combinatorial object of the form $(V,E,T)$ where $E \subseteq\binom{V}{2}$,
$T \subseteq\binom{V}{3}$, and $\{i,j\},\{j,k\},\{k,i\} \in E$ for all
$\{i,j,k\} \in T$ is called a $2$-dimensional simplicial complex. This is a
generalization of the notion of a graph, which is a $1$-dimensional simplicial
complex. In particular, given a graph $G= (V, E)$, the $2$-dimensional
simplicial complex $(V,E,T(E))$ is called the $3$-\textit{clique complex} of
$G$.

More generally, a \textit{simplicial complex} $(V,\Sigma)$ is a vertex set
$V=\{1,\dots,n\}$ together with a collection $\Sigma$ of subsets of $V$ that
is closed under inclusion, i.e.\ if $\tau\in\Sigma$ and $\sigma\subset\tau$,
then $\sigma\in\Sigma$. The elements in $\Sigma$ are called \textit{simplices}%
. For example, a $0$-simplex is just an element $i\in V$ (recall that we do
not distinguish between $\tbinom{V}{1}$ and $V$), a $1$-simplex is a pair
$\{i,j\}\in\binom{V}{2}$, a $2$-simplex is a triple $\{i,j,k\}\in\binom{V}{3}%
$, and so on. For $k\leq n$, a $k$-simplex is a $(k+1)$-element set in
$\binom{V}{k+1}$ and $\Sigma_{k}\subset\binom{V}{k+1}$ will denote the set of
all $k$-simplices in $\Sigma$. In the previous paragraph, $\Sigma_{0}=V$,
$\Sigma_{1}=E$, $\Sigma_{2}=T$, and $\Sigma=V\cup E\cup T$. In general, given
any undirected graph $G=(V,E)$, one obtains a $(k-1)$-dimensional simplicial
complex $K_{G}^{k}:=(V,\Sigma_{k-1})$ called the $k$-\textit{clique
complex}\footnote{Note that a $k$-clique is a $(k-1)$-simplex.} of $G$ by
`filling in' all its $j$-cliques for $j=1,\dots,k$, or more precisely, by
setting $\Sigma=\{j\text{-cliques of }G\mid j=1,\dots,k\}$. The $k$-clique
complex of $G$ where $k$ is maximal is just called the clique complex of $G$
and denoted $K_{G}$.

In this paper, we will mainly concern ourselves with studying the $3$-clique
complex $K_{G}^{3} = (V, E, T(E))$ where $G$ is a pairwise comparison graph.
Note that we could also look at the simplicial complex $(V,E,T_{\gamma}(E))$
where
\[
T_{\gamma}(E) := \bigl\{\{i,j,k\} \in T(E) \bigm| \lvert X_{ij} + X_{jk} +
X_{ki} \rvert\leq\gamma\bigr\}
\]
where $0 \le\gamma\le\infty$. For $\gamma= \infty$, we get $K_{G}^{3}$ but for
general $\gamma$ we get a subcomplex of $K_{G}^{3}$. We have found this to be
a useful multiscale characterization of the inconsistencies of pairwise
rankings but the detailed discussion will have to be left to a future paper.

\subsection{Cochains, Coboundary Maps, and Combinatorial Laplacians}

We will now introduce some discrete exterior calculus on a simplicial complex
where potential functions (scores or utility), edge flow (pairwise ranking),
triangular flow (triplewise ranking), gradient (global ranking induced by
scores), curl (local inconsistency) become just special cases of a much more
general framework. We will now also define the notions of
\textit{combinatorial divergence} and \textit{combinatorial Laplacians}. A
$0$-dimensional combinatorial Laplacian is just the usual graph Laplacian but
the case of greatest interest to us is the $1$-dimensional combinatorial
Laplacian, or what we will call the \textit{graph Helmholtzian}.

\begin{definition}
Let $K$ be a simplicial complex and recall that $\Sigma_{k}$ denotes its set
of $k$-simplices. A $k$-dimensional \textbf{cochain} is a real-valued function
on $k$-tuples of vertices that is alternating on each of the $k$-simplex and
$0$ otherwise, i.e.\ $f:V^{k}\rightarrow\mathbb{R}$ such that%
\[
f(i_{\sigma(0)},\dots,i_{\sigma(k)})=\operatorname*{sign}(\sigma)f(i_{0}%
,\dots,i_{k}),
\]
for all $(i_{0},\dots,i_{k})\in V^{k}$ and all $\sigma\in\mathfrak{S}_{k+1}$,
the permutation group on $k+1$ elements, and that%
\[
f(i_{0},\dots,i_{k})=0\quad\text{if }\{i_{0},\dots,i_{k}\}\notin\Sigma_{k}.
\]
The set of all $k$-cochains on $K$ is denoted $C^{k}(K,\mathbb{R})$.
\end{definition}

For simplicity we will often just write $C^{k}$ for $C^{k}(K,\mathbb{R})$. In
particular, $C^{0}$ is the space of potential functions (score/utility
functions), $C^{1}$ is the space of edge flows (pairwise rankings), and
$C^{2}$ is the space of triangular flows (triplewise rankings).

The $k$-cochain space $C^{k}$ can be given a choice of inner product. In view
of the weighted $l_{2}$-minimization for our statistical ranking problem
\eqref{eq:main2}, we will define the following inner product on $C^{1}$,
\begin{equation}
\langle X,Y\rangle_{w}=\sum\nolimits_{\{i,j\}\in E}w_{ij}X_{ij}Y_{ij},
\label{eq:ip1}%
\end{equation}
for all edge flows $X,Y\in C^{1}$. In the context of a pairwise comparison
graph $G$, it may not be immediately clear why this defines an inner product
since we have noted after \eqref{eq:W} that $W=[w_{ij}]$ is only a nonnegative
matrix and it is possible that some entries are $0$. However observe that by
definition $w_{ij}=0$ iff no voters have rated both alternatives $i$ and $j$
and therefore $\{i,j\}\not \in E$ by \eqref{eq:Edge} and so any edge flow $X$
will automatically have $X_{ij}=0$ by \eqref{eq:EdgeFlows}. Hence we indeed
have that $\langle X,X\rangle_{w}=0$ iff $X=0$, as required for an inner
product (the other properties are trivial to check).

The operators $\operatorname*{grad}$ and $\operatorname*{curl}$ are all
special instances of coboundary maps as defined below.

\begin{definition}
The $k$th \textbf{coboundary} operator $\delta_{k} : C^{k}(K, \mathbb{R}) \to
C^{k+1}(K,\mathbb{R})$ is the linear map that takes a $k$-cochain $f \in
C^{k}$ to a $(k+1)$-cochain $\delta_{k} f \in C^{k+1}$ defined by
\[
(\delta_{k} f)(i_{0},i_{1},\dots,i_{k+1}) := \sum\nolimits_{j=0}^{k+1}
(-1)^{j} f(i_{0},\dots,i_{j-1},i_{j+1},\dots,i_{k+1}).
\]

\end{definition}

Note that $i_{j}$ is omitted from $j$th term in the sum, i.e.\ coboundary maps
compute an alternating difference with one input left out. In particular,
$\delta_{0}=\operatorname*{grad}$, i.e.\ $(\delta_{0}s)(i,j)=s_{j}-s_{i}$, and
$\delta_{1}=\operatorname*{curl}$, i.e.\ $(\delta_{1}X)(i,j,k)=X_{ij}%
+X_{jk}+X_{ki}$.

Given a choice of an inner product $\langle\cdot, \cdot\rangle_{k}$ on $C^{k}%
$, we may define the adjoint operator of the coboundary map, $\delta_{k}^{*}:
C^{k+1} \to C^{k}$ in the usual manner, i.e.\ $\langle\delta_{k} f_{k},
g_{k+1} \rangle_{k+1} = \langle f_{k}, \delta_{k}^{*} g_{k+1}\rangle_{k}$.

\begin{definition}
The \textbf{combinatorial divergence} operator $\operatorname*{div} : C^{1}
(K, \mathbb{R}) \to C^{0} (K, \mathbb{R})$ is the adjoint of $\delta_{0} =
\operatorname*{grad}$, i.e.
\begin{equation}
\operatorname*{div} := - \delta_{0}^{*}.
\end{equation}

\end{definition}

Divergence will appear in the minimum norm solution to \eqref{eq:main2} and
can be used to characterize $\mathcal{M}_{G}^{\perp}$. As usual, we will drop
the adjective `combinatorial' when there is no cause for confusion.

For statistical ranking, it suffices to consider the cases $k=0,1,2$. Let $G$
be a pairwise comparison graph and $K_{G}$ its clique
complex\footnote{\label{foot:higher}It does not matter whether we consider
$K_{G}$ or $K_{G}^{3}$ or indeed any $K_{G}^{k}$ where $k\geq3$; the
higher-dimensional $k$-simplices where $k\geq3$ do not play a role in the
coboundary maps $\delta_{0},\delta_{1},\delta_{2}$.}. The cochain maps,
\begin{equation}
C^{0}(K_{G},\mathbb{R})\xrightarrow{\delta_0}C^{1}(K_{G},\mathbb{R}%
)\xrightarrow{\delta_1}C^{2}(K_{G},\mathbb{R}) \label{eq:cochainmap}%
\end{equation}
and their adjoint,
\begin{equation}
C^{0}(K_{G},\mathbb{R})\xleftarrow{\delta_0^*}C^{1}(K_{G},\mathbb{R}%
)\xleftarrow{\delta_1^*}C^{2}(K_{G},\mathbb{R}),
\end{equation}
have the following ranking theoretic interpretation with $C^{0},C^{1},C^{2}$
representing the spaces of score or utility functions, pairwise rankings, and
triplewise rankings respectively,
\begin{gather*}
\mathit{scores}\xrightarrow{\operatorname*{grad}}\mathit{pairwise}%
\xrightarrow{\operatorname*{curl}}\mathit{triplewise},\\
\mathit{scores}%
\xleftarrow{-\operatorname*{div}=\operatorname*{grad}^*}\mathit{pairwise}%
\xleftarrow{\operatorname*{curl}^*}\mathit{triplewise}.
\end{gather*}
In summary, the formulas for combinatorial gradient, curl, and divergence are
given by
\begin{align*}
&  (\operatorname*{grad}s)(i,j)=(\delta_{0}s)(i,j)=s_{j}-s_{i},\\
&  (\operatorname*{curl}X)(i,j,k)=(\delta_{1}X)(i,j,k)=X_{ij}+X_{jk}+X_{ki},\\
&  (\operatorname*{div}X)(i)=-(\delta_{0}^{\ast}X)(i)={\sum
\nolimits_{j\;\mathrm{s.t.}\,\{i,j\}\in E}}w_{ij}X_{ij}%
\end{align*}
with respect to the inner product $\langle X,Y\rangle_{w}=\sum_{\{i,j\}\in
E}w_{ij}X_{ij}Y_{ij}$ on $C^{1}$.

As an aside, it is perhaps worth pointing out that there is no special name
for the adjoint of $\operatorname*{curl}$ coming from physics because in
$3$-space, $C^{1}$ may be identified with $C^{2}$ via a property called
\textit{Hodge duality} and in which case $\operatorname*{curl}$ is a
self-adjoint operator, i.e.\ $\operatorname*{curl}^{*} = \operatorname*{curl}%
$. This will not be true in our case.

If we represent functions on vertices by $n$-vectors, edge flows by $n\times
n$ skew-symmetric matrices, and triangular flows by $n\times n\times n$
skew-symmetric hypermatrices, i.e.
\begin{align*}
C^{0}  &  =\mathbb{R}^{n},\\
C^{1}  &  =\{[X_{ij}]\in\mathbb{R}^{n\times n}\mid X_{ij}=-X_{ji}%
\}=\mathcal{A},\\
C^{2}  &  =\{[\Phi_{ijk}]\in\mathbb{R}^{n\times n\times n}\mid\Phi_{ijk}%
=\Phi_{jki}=\Phi_{kij}=-\Phi_{jik}=-\Phi_{ikj}=-\Phi_{kji}\},
\end{align*}
then in the language of linear algebra introduced in Section \ref{skew}, we
have the following correspondence
\begin{align*}
\operatorname*{im}(\delta_{0})  &  =\operatorname*{im}(\operatorname*{grad}%
)=\mathcal{M}_{G}, & \ker(\delta_{1})  &  =\ker(\operatorname*{curl}%
)=\mathcal{M}_{T},\\
\ker(\delta_{0}^{\ast})  &  =\ker(\operatorname*{div})=\mathcal{M}_{G}^{\perp
}, & \operatorname*{im}(\delta_{1}^{\ast})  &  =\operatorname*{im}%
(\operatorname*{curl}\nolimits^{\ast})=\mathcal{M}_{T}^{\perp},
\end{align*}
where $T=T(E)$.

Coboundary maps have the following important property.

\begin{lemma}
[Closedness]\label{lem:close} $\delta_{k+1}\circ\delta_{k} = 0$.
\end{lemma}

For $k=0$, this and its adjoint are well-known identities in vector calculus,
\begin{equation}
\operatorname*{curl}\circ\operatorname*{grad}=0,\qquad\operatorname*{div}%
\circ\operatorname*{curl}\nolimits^{\ast}=0. \label{eq:curlgrad0}%
\end{equation}
Ranking theoretically, the first identity simply says that a global ranking
must be consistent.

We will now define combinatorial Laplacians, higher-dimensional analogues of
the graph Laplacian.

\begin{definition}
Let $K$ be a simplicial complex. The $k$-dimensional \textbf{combinatorial
Laplacian} is the operator $\Delta_{k}: C^{k}(K,\mathbb{R})\to C^{k}%
(K,\mathbb{R})$ defined by
\begin{equation}
\label{eq:laplacian}\Delta_{k} = \delta_{k}^{*}\circ\delta_{k} + \delta_{k-1}
\circ\delta_{k-1}^{*}.
\end{equation}

\end{definition}

In particular, for $k=0$,
\[
\Delta_{0}=\delta_{0}^{\ast}\circ\delta_{0}=\operatorname*{div}\circ
\operatorname*{grad}%
\]
is a discrete analogue of the scalar Laplacian or Laplace operator while for
$k=1$,
\[
\Delta_{1}=\delta_{1}^{\ast}\circ\delta_{1}+\delta_{0}\circ\delta_{0}^{\ast
}=\operatorname*{curl}\nolimits^{\ast}\circ\operatorname*{curl}%
-\operatorname*{grad}\circ\operatorname*{div}%
\]
is a discrete analogue of the vector Laplacian or Helmholtz operator. In the
context of graph theory, if $K=K_{G}$, then $\Delta_{0}$ is called the graph
Laplacian \cite{Chung97} while $\Delta_{1}$ is called the \textit{graph
Helmholtzian}.

The combinatorial Laplacian has some well-known, important properties.

\begin{lemma}
\label{lem:betti} $\Delta_{k}$ is a positive semidefinite operator.
Furthermore, the dimension of $\ker(\Delta_{k})$ is equal to $k$th Betti
number of $K$.
\end{lemma}

We will call a cochain $f \in\ker(\Delta_{k})$ \textit{harmonic} since they
are solutions to higher-dimensional analogue of the Laplace equation
\[
\Delta_{k} f = 0.
\]
Strictly speaking, the Laplace equation refers to $\Delta_{0} f = 0$. The
equation $\Delta_{1} X = 0$ is really the \textit{Helmholtz equation}. But
nonetheless, we will still call an edge flow $X \in\ker(\Delta_{1})$ a
harmonic flow.

\subsection{Hodge Decomposition Theorem}

We now state the main theorem in combinatorial Hodge theory.

\begin{theorem}
[Hodge Decomposition Theorem]\label{thm:hodge} $C^{k}(K,\mathbb{R})$ admits an
orthogonal decomposition
\[
C^{k}(K,\mathbb{R}) = \operatorname*{im} (\delta_{k-1}) \oplus\ker(\Delta_{k})
\oplus\operatorname*{im}(\delta_{k}^{*}).
\]
Furthermore,
\[
\ker(\Delta_{k}) = \ker(\delta_{k}) \cap\ker(\delta_{k-1}^{*}).
\]

\end{theorem}

An elementary proof targeted at a computer science readership may be found in
\cite{Friedman96}. For completeness we include a proof here.

\begin{proof}
We will use Lemma \ref{lem:close}. First, $C^{k}=\operatorname*{im}%
(\delta_{k-1})\oplus\ker(\delta_{k-1}^{\ast})$. Since $\delta_{k}\delta
_{k-1}=0$, taking adjoint yields $\delta_{k-1}^{\ast}\delta_{k}^{\ast}=0$,
which implies that $\operatorname*{im}(\delta_{k}^{\ast})\subseteq\ker
(\delta_{k-1}^{\ast})$. Therefore $\ker(\delta_{k-1}^{\ast}%
)=[\operatorname*{im}(\delta_{k}^{\ast})\oplus\ker(\delta_{k})]\cap\ker
(\delta_{k-1}^{\ast})=[\operatorname*{im}(\delta_{k}^{\ast})\cap\ker
(\delta_{k-1}^{\ast})]\oplus\lbrack\ker(\delta_{k})\cap\ker(\delta_{k-1}%
^{\ast})]=\operatorname*{im}(\delta_{k}^{\ast})\oplus\lbrack\ker(\delta
_{k})\cap\ker(\delta_{k-1}^{\ast})]$. It remains to show that $\ker(\delta
_{k})\cap\ker(\delta_{k-1}^{\ast})=\ker(\Delta_{k})=\ker(\delta_{k-1}%
\delta_{k-1}^{\ast}+\delta_{k}^{\ast}\delta_{k})$. Clearly $\ker(\delta
_{k})\cap\ker(\delta_{k-1}^{\ast})\subseteq\ker(\Delta_{k})$. For any
$X=\delta_{k}^{\ast}\Phi\in\operatorname*{im}(\delta_{k}^{\ast})$ where
$0\neq\Phi\in C^{k+1}$, Lemma \ref{lem:close} again implies $\delta
_{k-1}\delta_{k-1}^{\ast}X=\delta_{k-1}\delta_{k-1}^{\ast}\delta_{k}^{\ast
}\Phi=0$, but $\delta_{k}^{\ast}\delta_{k}X=\delta_{k}^{\ast}\delta_{k}%
\delta_{k}^{\ast}\Phi\neq0$, which implies that $\Delta_{k}X\neq0$. Similarly
for $X\in\operatorname*{im}(\delta_{0})$. Hence $\ker(\Delta_{k})=\ker
(\delta_{k})\cap\ker(\delta_{k-1}^{\ast})$.
\end{proof}

While Hodge decomposition holds in general for any simplicial complex and in
any dimension $k$, the case $k = 1$ is more often called the \textit{Helmholtz
decomposition theorem}\footnote{On a simply connected manifold, the continuous
version of the Helmholtz decomposition theorem is just the fundamental theorem
of vector calculus.}. We will state it here for the special case of a clique complex.

\begin{theorem}
[Helmholtz Decomposition Theorem]\label{thm:helmholz} Let $G=(V,E)$ be an
undirected, unweighted graph and $K_{G}$ be its clique complex. The space of
edge flows on $G$, i.e.\ $C^{1}(K_{G},\mathbb{R})$, admits an orthogonal
decomposition
\begin{align}
C^{1}(K_{G},\mathbb{R})  &  =\operatorname*{im}(\delta_{0})\oplus\ker
(\Delta_{1})\oplus\operatorname*{im}(\delta_{1}^{\ast}%
)\nonumber\label{eq:hodge}\\
&  =\operatorname*{im}(\operatorname*{grad})\oplus\ker(\Delta_{1}%
)\oplus\operatorname*{im}(\operatorname*{curl}\nolimits^{\ast}).
\end{align}
Furthermore,
\begin{equation}
\ker(\Delta_{1})=\ker(\delta_{1})\cap\ker(\delta_{0}^{\ast})=\ker
(\operatorname*{curl})\cap\ker(\operatorname*{div}). \label{eq:hodge2}%
\end{equation}

\end{theorem}

\begin{figure}[ptb]
\centering
\label{fig:hodge}
\resizebox{8cm}{!}{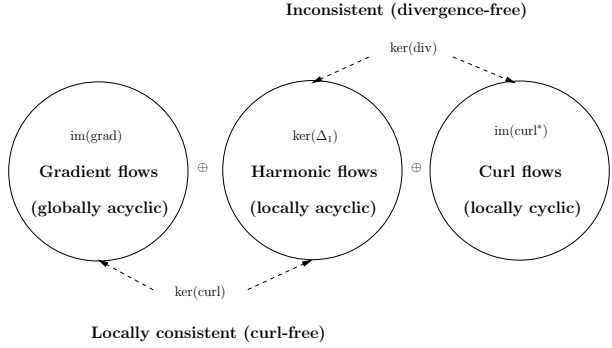}\caption{Hodge/Helmholtz
decomposition of pairwise rankings}%
\end{figure}

The clique complex $K_{G}$ above may be substituted with any $K_{G}^{k}$ with
$k \ge3$ (see Footnote \ref{foot:higher}). The equation \eqref{eq:hodge2} says
that an edge flow is harmonic iff it is both curl-free and divergence-free.
Figure \ref{fig:hodge} illustrates \eqref{eq:hodge}. 

To understand the significance of this theorem, we need to discuss the ranking
theoretic interpretations of each subspace in the theorem.

\begin{enumerate}
\item $\operatorname*{im} (\delta_{0})=\operatorname*{im}
(\operatorname*{grad})$ denotes the subspace of pairwise rankings that are the
gradient flows of score functions. Thus this subspace comprises the
\textit{globally consistent} or \textit{acyclic} pairwise rankings. Given any
pairwise ranking from this subspace, we may determine a score function on the
alternatives that is unique up to an additive
constant\footnote{\label{foot:gradker}Note that $\ker(\delta_{0}) =
\ker(\operatorname*{grad})$ is the set of constant functions on $V$ and so
$\operatorname*{grad}(s) = \operatorname*{grad} (s + \text{constant})$.} and
then we may rank all alternatives globally in terms of their scores.

\item $\ker(\delta_{0}^{*}) =\ker(\operatorname*{div})$ denotes the subspace
of \textit{divergence-free} pairwise rankings, whose total in-flow equals
total out-flow for each alternative $i\in V$. Such pairwise rankings may be
regarded as \textit{cyclic} rankings, i.e.\ rankings of the form $i \succeq j
\succeq k \succeq\dots\succeq i$, and they are clearly \textit{inconsistent}.
Since $\ker(\operatorname*{div}^{*}) = \operatorname*{im}(\operatorname*{grad}%
)^{\perp}$, cyclic rankings have zero projection on global rankings.

\item $\ker(\delta_{1})=\ker(\operatorname*{curl})$ denotes the subspace of
\textit{curl-free} pairwise rankings with zero flow-sum along any triangle in
$K_{G}$. This corresponds to \textit{locally consistent} (i.e.\ triangularly
consistent) pairwise rankings. Note that by the Closedness Lemma
$\operatorname*{curl}\circ\operatorname*{grad}=0$ and so $\operatorname*{im}%
(\operatorname*{grad}) \subseteq\ker(\operatorname*{curl})$. In general, the
globally consistent pairwise rankings induced by gradient flows of score
functions only account for a subset of locally consistent rankings. The
remaining ones are the locally consistent rankings that are not globally
consistent and they are precisely the harmonic rankings discussed below.

\item $\ker(\Delta_{1}) = \ker(\operatorname*{curl}) \cap\ker
(\operatorname*{div})$ denotes the subspace of \textit{harmonic} pairwise
rankings, or just harmonic rankings in short. It is the space of solutions to
the Helmholtz equation. Harmonic rankings are exactly those pairwise rankings
that are both \textit{curl-free} and \textit{divergence-free}. These are only
locally consistent with zero curl on every triangle in $T(E)$ but are not
globally consistent. In other words, while there are no inconsistencies due to
small loops of length $3$, i.e.\ $i \succeq j \succeq k \succeq i$, there are
inconsistencies along larger loops of lengths $> 3$, i.e.\ $a \succeq b
\succeq c \succeq\dots\succeq z \succeq a$. So these are also cyclic rankings.
Rank aggregation on $\ker(\Delta_{1})$ depends on the edge paths traversed in
the simplicial complex; along homotopy equivalent paths one obtains consistent
rankings. Figure \ref{fig:harmonic} gives an example of harmonic rankings.

\item $\operatorname*{im}(\delta_{1}^{\ast})=\operatorname*{im}%
(\operatorname*{curl}^{\ast})$ denotes the subspace of \textit{locally cyclic}
pairwise rankings that have non-zero curls along triangles. By the Closedness
Lemma, $\operatorname*{im}(\operatorname*{curl}^{\ast})\subseteq
\ker(\operatorname*{div})$ and so this subspace is in general a proper
subspace of the divergence-free rankings; the orthogonal complement of
$\operatorname*{im}(\operatorname*{curl}^{\ast})$ in $\ker(\operatorname*{div}%
)$ is precisely the space of harmonic rankings $\ker(\Delta_{1})$ discussed above.
\end{enumerate}

\section{Implications of Hodge Theory\label{imply}}

We now state two immediate implications of the Helmholtz decomposition theorem
when applied to statistical ranking. The first implication is that it gives an
interpretation of the solution and residual of the optimization problem
\eqref{eq:main2}; these are respectively the $l_{2}$-projection on gradient
flows and divergence-free flows. In the context of statistical ranking and in
the $l_{2}$-sense, the solution to \eqref{eq:main2} gives the nearest globally
consistent pairwise ranking to the data while the residual gives the sum total
of all inconsistent components (both local and harmonic) in the data. The
second implication is the condition that local consistency guarantees global
consistency whenever there is no harmonic component in the data (which happens
iff the clique complex of the pairwise comparison graph is `loop-free').

\subsection{Structure Theorem for Global Ranking and the Residual of
Inconsistency\label{structure}}

In order to cast our optimization problem \eqref{eq:main2} in the Hodge
theoretic framework, we need to specify relevant inner products on $C^{0},
C^{1}, C^{2}$. As before, the inner product on the space of edge flows
(pairwise rankings) $C^{1}$ will be a weighted Euclidean inner product
\[
\langle X,Y\rangle_{w} = \sum\nolimits_{\{i,j\}\in E} w_{ij} X_{ij} Y_{ij}
\]
for $X, Y \in C^{1}$. We will let the inner products on $C^{0}$ and $C^{2}$ be
the unweighted Euclidean inner product
\[
\langle r,s\rangle= \sum\nolimits_{i=1}^{n} r_{i} s_{i}, \qquad\langle
\Theta,\Phi\rangle= \sum\nolimits_{\{i,j,k\} \in T(E)} \Theta_{ijk}
\Phi_{ijk}
\]
for $r, s \in C^{0}$ and $\Theta, \Phi\in C^{2}$. We note that other inner
products can be chosen (e.g.\ the inner products on $C^{0}$ and $C^{2}$ could
have been weighted) with corresponding straightforward modification of
\eqref{eq:main2} but this would not change the essential nature of our
methods. We made the above choices mainly to keep our notations uncluttered.

The optimization problem \eqref{eq:main2} is then equivalent to an $l_{2}%
$-projection of an edge flow representing a pairwise ranking onto
$\operatorname*{im}(\operatorname*{grad})$,
\[
\min_{s \in C^{0}} \| \delta_{0} s - \bar{Y} \|_{2,w} = \min_{s \in C^{0}} \|
\operatorname*{grad} s - \bar{Y} \|_{2,w},
\]
The Helmholtz decomposition theorem then leads to the following result about
the structures of the solutions and residuals of \eqref{eq:main2}. In Theorem
\ref{thm:l2} below, we assume that the pairwise ranking data $\bar{Y}$ has
been estimated from one of the methods in Section \ref{exmp:netflix}. The
least squares solution $s$ will be a score function that induces
$\operatorname*{grad} s$, the $l_{2}$-nearest global ranking to $\bar{Y}$.
Since $s$ is only unique up to a constant (see Footnote \ref{foot:gradker}),
we determine a unique minimum norm solution $s^{*}$ for the sake of
well-posedness; but nevertheless any $s$ will yield the same global ordering
of alternatives. The least squares residual $R^{*}$ represents the
inconsistent component of the ranking data $\bar{Y}$. The magnitude of $R^{*}$
is a `certificate of reliability' for $s$; since if this is small, then the
globally consistent component $\operatorname*{grad} s$ accounts for most of
the variation in $\bar{Y}$ and we may conclude that $s$ gives a reasonably
reliable ranking of the alternatives. But even when the magnitude of $R^{*}$
is large, we will see that it may be further resolved into a global and a
local component that determine when a comparison of alternatives with respect
to $s$ is still valid.

\begin{theorem}
\label{thm:l2}

\begin{enumerate}
[(i)]

\item \label{A} Solutions of \eqref{eq:main2} satisfy the following normal
equation
\begin{equation}
\Delta_{0}s=-\operatorname*{div}\bar{Y}, \label{eq:normal}%
\end{equation}
and thus the minimum norm solution is
\begin{equation}
s^{\ast}=-\Delta_{0}^{\dag}\operatorname*{div}\bar{Y} \label{eq:glbrank}%
\end{equation}
where $\dag$ indicates a Moore-Penrose inverse. The divergence in
\eqref{eq:glbrank} is given by
\[
(\operatorname*{div}\bar{Y})(i)=\sum\nolimits_{j\;\mathrm{s.t.}\,\{i,j\}\in
E}w_{ij}\bar{Y}_{ij},
\]
and the matrix representing the graph Laplacian is given by
\[
\lbrack\Delta_{0}]_{ij}=%
\begin{cases}
\sum_{i}w_{ii} & \text{if }j=i,\\
-w_{ij} & \text{if }j\text{ is such that }\{i,j\}\in E,\\
0 & \text{otherwise.}%
\end{cases}
\]

\item \label{B} The residual $R^{\ast}=\bar{Y}-\delta_{0}s^{\ast}$ is
divergence-free, i.e.\ $\operatorname*{div}R^{\ast}=0$. Moreover, it has a
further orthogonal decomposition
\begin{equation}
R^{\ast}=\operatorname*{proj}\nolimits_{\operatorname*{im}%
(\operatorname*{curl}^{\ast})}\bar{Y}+\operatorname*{proj}\nolimits_{\ker
(\Delta_{1})}\bar{Y}, \label{eq:curl-free}%
\end{equation}
where $\operatorname*{proj}_{\operatorname*{im}(\operatorname*{curl}^{\ast}%
)}\bar{Y}$ is a local cyclic ranking accounting for local inconsistencies and
$\operatorname*{proj}_{\ker(\Delta_{1})}\bar{Y}$ is a harmonic ranking
accounting for global inconsistencies. In particular, the projections are
given by
\begin{equation}
\operatorname*{proj}\nolimits_{\operatorname*{im}(\operatorname*{curl}^{\ast
})}=\operatorname{curl}^{\dag}\operatorname{curl}\quad\text{and}%
\quad\operatorname*{proj}\nolimits_{\ker(\Delta_{1})}=I-\Delta_{1}^{+}%
\Delta_{1} \label{eq:4curls}%
\end{equation}

\end{enumerate}
\end{theorem}

\begin{proof}
The normal equation for the least squares problem $\min_{s\in C^{0}}%
\Vert\delta_{0}s-\bar{Y}\Vert_{2,w}^{2}$ is%
\[
\delta_{0}^{\ast}\delta_{0}s=\delta_{0}^{\ast}\bar{Y}.
\]
\eqref{eq:normal}, \eqref{eq:glbrank}, and $\operatorname{div}R^{\ast}=0$ are
obvious upon substituting $\Delta_{0}=\delta_{0}^{\ast}\delta_{0}$ and
$\operatorname*{div}=-\delta_{0}^{\ast}$. The expressions for divergence and
graph Laplacian in (\ref{A}) follow from their respective definitions. The
Helmholtz decomposition theorem implies%
\[
\ker(\Delta_{1})\oplus\operatorname*{im}(\operatorname{curl}^{\ast
})=\operatorname*{im}(\operatorname{grad})^{\perp}.
\]
Obviously $\operatorname*{proj}\nolimits_{\operatorname*{im}%
(\operatorname{grad})^{\perp}}\operatorname{grad}s^{\ast}=0$. Since $R^{\ast
}=\bar{Y}-\operatorname{grad}s^{\ast}$ is a least squares residual, we must
have $\operatorname*{proj}\nolimits_{\operatorname*{im}(\operatorname{grad}%
)}R^{\ast}=\operatorname*{proj}%
\nolimits_{\operatorname*{im}(\operatorname{grad})}\bar{Y}-\operatorname{grad}s^{\ast}=0$. These
observations yield \eqref{eq:curl-free}, as%
\[
R^{\ast}=\operatorname*{proj}\nolimits_{\operatorname*{im}(\operatorname{grad}%
)}R^{\ast}+\operatorname*{proj}\nolimits_{\operatorname*{im}%
(\operatorname{grad})^{\perp}}R^{\ast}=0+\operatorname*{proj}\nolimits_{\ker
(\Delta_{1})\oplus\operatorname*{im}(\operatorname{curl}^{\ast})}\bar{Y}.
\]
The expression for the projection in \eqref{eq:4curls} is standard.
\end{proof}

In the special case when the pairwise ranking matrix $G$ is a complete graph
and we have an unweighted Euclidean inner product on $C^{1}$, the minimum norm
solution $s^{*}$ in \eqref{eq:glbrank} satisfies $\sum_{i} s^{*}_{i} = 0$ and
is given by
\begin{equation}
\label{eq:borda0}s^{*}_{i} = - \frac{1}{n} \operatorname*{div}(\bar{Y})(i) = -
\frac{1}{n} \sum\nolimits_{j} \bar{Y}_{ij}.
\end{equation}
In Section \ref{social}, we shall see that this is the well-known
\textit{Borda count} in social choice theory, a measure that is also widely
used in psychology and statistics
\cite{KenSmi40,Mosteller51a,Mosteller51b,Mosteller51c,David69}. Since $G$ is a
complete graph only when the ranking data is complete, i.e.\ every voter has
rated every alternative, this is an unrealistic scenario for the type of
modern ranking data discussed in Section \ref{intro}. Among other things, the
Hodge theoretic framework generalizes Borda count to scenarios where the
ranking data is incomplete or even highly incomplete.

In (\ref{B}) the locally cyclic ranking component is obtained by solving
\[
\min_{\Phi\in C^{2}}\Vert\operatorname*{curl}\nolimits^{\ast}\Phi-R^{\ast
}\Vert_{2,w}=\min_{\Phi\in C^{2}}\Vert\operatorname*{curl}\nolimits^{\ast}%
\Phi-\bar{Y}\Vert_{2,w}.
\]
The above equality implies that there is no need to first solve for $R^{\ast}$
before we may obtain $\Phi$; one could get it directly from the pairwise
ranking data $\bar{Y}$. Note that the solution is only determined up to an
additive term of the form $\operatorname*{grad}s$ since by virtue of
\eqref{eq:curlgrad0},
\begin{equation}
\operatorname*{curl}(\Phi+\operatorname*{grad}s)=\operatorname*{curl}\Phi.
\label{eq:indet1}%
\end{equation}
For the sake of well-posedness, we will seek the unique minimum norm solution
given by
\[
\Phi^{\ast}=(\delta_{1}\circ\delta_{1}^{\ast})^{\dag}\delta_{1}\bar
{Y}=(\operatorname*{curl}\circ\operatorname*{curl}\nolimits^{\ast})^{\dag
}\operatorname*{curl}\bar{Y}%
\]
and the required component is given by $\operatorname*{proj}%
_{\operatorname*{im}(\operatorname*{curl}^{\ast})}\bar{Y}=\operatorname*{curl}%
^{\ast}\Phi^{\ast}$. The reader may have noted a parallel between the two
problems
\[
\min_{s\in C^{0}}\lVert\operatorname*{grad}s-\bar{Y}\rVert_{2,w}%
\qquad\text{and}\qquad\min_{\Phi\in C^{2}}\lVert\operatorname*{curl}%
\nolimits^{\ast}\Phi-\bar{Y}\rVert_{2,w}.
\]
Indeed in many contexts, $s$ is called the \textit{scalar potential} while
$\Phi$ is called the \textit{vector potential}. As seen earlier in Definition
\ref{def:grad}, an edge flow of the form $\operatorname*{grad}s$ for some
$s\in C^{0}$ is called a gradient flow; in analogy, we will call an edge flow
of the form $\operatorname*{curl}^{\ast}\Phi$ for some $\Phi\in C^{2}$ a
\textit{curl flow}.

We note that the $l_{2}$-residual $R^{\ast}$, being divergence-free, is a
cyclic ranking. Much like \eqref{eq:indet1}, the divergence-free condition is
satisfied by a whole family of edge flows that differs from $R^{\ast}$ only by
a term of the form $\operatorname*{curl}^{\ast}\Phi$ since
\[
\operatorname*{div}(R^{\ast}+\operatorname*{curl}\nolimits^{\ast}%
\Phi)=\operatorname*{div}R^{\ast}%
\]
because of \eqref{eq:curlgrad0}. The subset of $C^{1}$ given by
\[
\{R^{\ast}+\operatorname*{curl}\nolimits^{\ast}\Phi\mid\Phi\in C^{2}\}
\]
is called the \textit{homology class} of $R^{\ast}$. The harmonic ranking
$\operatorname*{proj}_{\ker(\Delta_{1})}\bar{Y}$ is just one element in this
class\footnote{Two elements of the same homology class are called
\textit{homologous}.}. In general, it will be dense in the sense that it will
be nonzero on almost every edge in $E$. This is because in addition to the
divergence-free condition, the harmonic ranking must also satisfy the
curl-free condition by virtue of \eqref{eq:hodge2}. So if parsimony or
sparsity is the objective, e.g.\ if one wants to identify a small number of
conflicting comparisons that give rise to the inconsistencies in the ranking
data, then the harmonic ranking does not offer much information in this
regard. To better understand ranking inconsistencies via the structure of
$R^{\ast}$, it is often helpful to look for elements in the same homology
class with the \textit{sparsest support}, i.e.
\[
\min_{\Phi\in C^{2}}\lVert\operatorname*{curl}\nolimits^{\ast}\Phi-R^{\ast
}\rVert_{0}=\min_{\Phi\in C^{2}}\lVert\operatorname*{curl}\nolimits^{\ast}%
\Phi-\operatorname*{proj}\nolimits_{\ker(\Delta_{1})}\bar{Y}\rVert_{0}.
\]

The widely used convex relaxation replacing the $l_{0}$-`norm' by the $l_{1}%
$-norm may be employed \cite{Ali08}, i.e.
\[
\min_{\Phi\in C^{2}}\lVert\operatorname*{curl}\nolimits^{\ast}\Phi-R^{\ast
}\rVert_{1}:=\min_{\Phi\in C^{2}}\sum\nolimits_{i,j}\lvert
(\operatorname*{curl}\nolimits^{\ast}\Phi)_{ij}-R_{ij}^{\ast}\rvert.
\]
A solution $\tilde{\Phi}$ of such an $l_{1}$-minimization problem is expected
to give a sparse element $R^{\ast}-\operatorname*{curl}^{\ast}\tilde{\Phi}$,
which we call an $l_{1}$-\textit{approximate sparse generator} of $R^{\ast}$,
or equivalently, of $\operatorname*{proj}_{\ker(\Delta_{1})}\bar{Y}$. We will
discuss them in detail in Section \ref{sparse}. The bottom line here is that
we want to find the shortest cycles that represent the global inconsistencies
and perhaps remove the corresponding edges in the pairwise comparison graph,
in view of what we will discuss next in Section \ref{local-global}. One
plausible strategy to get a globally consistent ranking is to remove a number
of problematic `conflicting' comparisons from the pairwise comparison graph.
Since it is only reasonable to remove as few edges as possible, this
translates to finding a homology class with the sparsest support. This is
similar to the minimum feedback arc set approach discussed in Section
\ref{compare}.

We will end the discussion of this section with a note on computational
costs. Solving for a global ranking $s^{*}$ in \eqref{eq:glbrank} only
requires the solution of an $n\times n$ least squares problem, which comes
with a modest cost of $O(n^{3})$ flops ($n = \lvert V \rvert $). As we
note later in Section~\ref{exp:pagerank}, for web ranking analysis such a cost is
no more than computing the PageRank. On the other hand, the analysis of
inconsistency is generally harder. For example, evaluating curls requires
$\lvert T \rvert $ flops and this is $\binom{n}{3} \sim O(n^3)$ in the
worst case. Since an actual computation of $\Phi^{*}$ involves solving a
least squares problem of size $\lvert T \rvert \times \lvert T \rvert $,
the computation cost incurred is of order $O(n^{9})$. Nevertheless, any
sparsity in the data (when $\lvert T \rvert \ll n^3$) may be exploited by
choosing the right least squares solver. For example, one may use the
general sparse least squares solver \textsc{lsqr} \cite{LSQR} or the new
\textsc{minres}-\textsc{qlp} \cite{Choi06,ChoiSaunders} that works
specifically for symmetric matrices. We will leave discussions of actual
computations and more extensive numerical experiments to a future article.
It suffices to note here that it is in general harder to isolate the
harmonic component of the ranking data than the globally consistent
component.

\subsection{Local Consistency versus Global Consistency\label{local-global}}

In this section, we discuss a useful result, that local consistency implies
global consistency whenever the harmonic component is absent from the ranking
data. Whether a harmonic component exists is dependent on the topology of the
clique complex $K_{G}^{3}$. We will invoke the recent work of Kahle
\cite{Kahle06} on such topological properties of random graphs to argue that
harmonic components are exceedingly unlikely to occur.

By Lemma \ref{lem:betti}, the dimension of $\ker(\Delta_{1})$ is equal to the
first Betti number $\beta_{1}(K)$ of the underlying simplicial complex $K$. In
particular, we know that $\ker(\Delta_{1})=0$ if $\beta_{1} (K)=0$, and so the
harmonic component of any edge flow on $K$ is automatically absent when
$\beta_{1} (K)=0$ (roughly speaking, $\beta_{1} (K)=0$ means that $K$ does not
have any $1$-dimensional holes). This leads to the following result.

\begin{theorem}
\label{thm:consistency} Let $K_{G}^{3} = (V, E, T(E))$ be a $3$-clique complex
of a pairwise comparison graph $G = (V, E)$. If $K_{G}^{3}$ does not contain
any $1$-loops, i.e.\ $\beta_{1}(K_{G}^{3})=0$, then every locally consistent
pairwise ranking is also globally consistent. In other words, if the edge flow
$X \in C^{1}(K_{G}^{3},\mathbb{R})$ is curl-free, i.e.
\[
\operatorname*{curl}(X) (i,j,k) = 0
\]
for all $\{ i,j,k \} \in T(E)$, then it is a gradient flow, i.e.\ there exists
$s \in C^{0}(K_{G}, \mathbb{R})$ such that
\[
X = \operatorname*{grad} s.
\]

\end{theorem}

\begin{proof}
This follows from the Helmholtz decomposition theorem since $\dim(\ker
\Delta_{1})=\beta_{1}(K_{G}^{3})=0$ and so any $X$ that is curl-free is
automatically in $\operatorname*{im}(\operatorname*{grad})$.
\end{proof}

When $G$ is a complete graph, then we always have that $\beta_{1}(K_{G})=
\beta_{1}(K_{G}^{3})=0$ and this justifies the discussion after Definition
\ref{def:consistency} about the equivalence of local and global consistencies
for complete pairwise comparison graphs. In general, $G$ will be incomplete
due to missing ranking data (not all voters have rated all alternatives) but
as long as $K_{G}^{3}$ is \textit{loop-free}, such a claim still holds. In
finance, this theorem translates into the well-known result that ``triangular
arbitrage-free implies arbitrage-free.'' The theorem enables us to infer
global consistency from a local condition --- whether the ranking data is
curl-free. We note that being curl-free is a strong condition. If we instead
have ``triangular transitivity'' in the ordinal sense, i.e.\ $a \succeq b
\succeq c$ implies $a \succeq c$, then there is no result analogous to Theorem
\ref{thm:consistency}.

At least for Erd\"{o}s-R\'{e}nyi random graphs, the Betti number $\beta_{1}$
could only be non-zero when the edges are neither too sparse nor too dense.
The following result by Kahle \cite{Kahle06} quantifies this statement. He
showed that $\beta_{1}$ undergoes two phase transitions from zero to nonzero
and back to zero as the density of edges grows.

\begin{theorem}
[Kahle 2006]\label{thm:kahle} For an Erd\"{o}s-R\'{e}nyi random graph $G(n,p)$
on $n$ vertices where the edges are independently generated with probability
$p$, its clique complex $K_{G}$ almost always has $\beta_{1}(K_{G}) = 0$,
except when
\begin{equation}
\label{eq:kahle}\frac{1}{n^{2}} \ll p \ll\frac{1}{n}.
\end{equation}

\end{theorem}

Without getting into a discussion about whether Erd\"{o}s-R\'{e}nyi random
graphs are good models for pairwise ranking comparison graphs of real-world
ranking data, we note that the Netflix pairwise comparison graph has a high
probability of having $\beta_{1}(K_{G}) = 0$ if Kahle's result applies.
Although the original customer-product rating matrix of the Netflix prize
dataset is highly incomplete (more than $99\%$ missing values), its pairwise
comparison graph is very dense (less than $0.22\%$ missing edges). In other
words, $p$ (probability of an edge) and $n$ (number of vertices) are both
large and so \eqref{eq:kahle} is not satisfied.

\section{$l_{1}$-aspects of Hodge Theoretic Ranking\label{L1}}

Hodge theory is by and large an $l_{2}$-theory: inner products on cochains,
adjoint of coboundary operators, orthogonality of Hodge decomposition, are all
naturally associated with (weighted or unweighted) $l_{2}$-norms. In this
section, we will take an oblique approach and study the $l_{1}$-aspects of
combinatorial Hodge theory in the context of statistical ranking, with
robustness and parsimony (or sparsity) being our two obvious motivations. We
will study two $l_{1}$-norm minimization problems: (1) the $l_{1}$-projection
on gradient flows (globally consistent rankings), which we show to have a dual
problem as correlation maximization over bounded divergence-free flows (cyclic
rankings); (2) an $l_{1}$-approximation to find sparse divergence-free flows
(cyclic rankings) homologous to the residual of the $l_{2}$-projection, which
we show to have a dual problem as correlation maximization over bounded
curl-free flows (locally consistent rankings). We observe that the primal
versus dual relation is revealed as an `$\operatorname*{im}%
(\operatorname*{grad})$ versus $\ker(\operatorname*{div})$' relation in first
case and an `$\operatorname*{im}(\operatorname*{curl}^{\ast})$ versus
$\ker(\operatorname*{curl})$' relation in the second case.

\subsection{Robust Ranking: $l_{1}$-projection on gradient flows\label{robust}%
}

We have briefly mentioned this problem in Section \ref{main} as an $l_{1}%
$-variation of the least squares model \eqref{eq:main2} for statistical
ranking. Here we will derive a duality result for \eqref{eq:mainl1}. As
before, we assume a pairwise comparison graph $G=(V,E)$ and an edge flow
$\bar{Y}\in C^{1}(K_{G},\mathbb{R})$ that comes from our ranking data.
Consider the following minimization problem,
\begin{equation}%
\begin{array}
[c]{rcl}%
\min &  & \Vert X-\bar{Y}\Vert_{1,w}\\
\mathrm{s.t.} &  & X=\operatorname*{grad}s,\\
&  & X=-X^{\top},
\end{array}
\label{eq:l1}%
\end{equation}
which may be regarded as the $l_{1}$-projection\footnote{The projection of a
point $X$ onto a closed subset $S$ in a finite-dimensional norm space is
simply the unique point $X_{S}\in S$ that is nearest to $X$ in the norm.} of
an edge flow $\bar{Y}$ onto the space of gradient flows,
\begin{equation}
\min_{s\in C^{0}}\lVert\operatorname*{grad}s-\bar{Y}\rVert_{1,w}=\min_{s\in
C^{0}}\sum\nolimits_{\{i,j\}\in E}w_{ij}|s_{j}-s_{i}-\bar{Y}_{ij}|.
\label{eq:l1constant}%
\end{equation}
In other words, we attempt to find the nearest globally consistent ranking
$\operatorname*{grad}s$ to the pairwise ranking $\bar{Y}$ as measured by the
$l_{1}$-norm. Such a norm is often employed in robust regression since its
solutions will be relatively more robust to outliers or large deviations in
the ranking data $\bar{Y}$ when compared to the $l_{2}$-norm in
\eqref{eq:main2} \cite{NarWel82,CorMohRas07}. The computational cost paid in
going from \eqref{eq:main2} to \eqref{eq:l1} is that of replacing a linear
least squares problem with a linear programming problem.

Recall that the minimum norm $l_{2}$-minimizer is given by $s^{\ast}%
=-(\Delta_{0})^{\dag}\operatorname*{div}\bar{Y}$ and the $l_{2}$-residual is
given by $R^{\ast}=\bar{Y}-\operatorname*{grad}s^{\ast}$. Hence
\[
\min_{s\in C^{0}}\lVert\operatorname*{grad}s-\bar{Y}\rVert_{1,w}%
=\min_{s^{\prime}\in C^0}\lVert\operatorname*{grad}s^{\prime}-R^*\rVert_{1,w}%
\]
where $s^{\prime} = s - s^*$. It follows that the $l_{1}$-minimizers in
\eqref{eq:l1constant} may be characterized by\footnote{Recall that
$\operatorname*{argmin}$ refers to the \textit{set} of all minimizers. The
addition of sets here is just the usual Minkowski sum.}%
\begin{multline*}
\operatorname*{argmin}\nolimits_{s\in C^{0}}\Vert\operatorname*{grad}s-\bar
{Y}\Vert_{1,w}=\operatorname*{argmin}\nolimits_{s\in C^{0}}\Vert
\operatorname*{grad}s-\bar{Y}\Vert_{2,w}\\
+\operatorname*{argmin}\nolimits_{s^{\prime}\in C^0}\Vert\operatorname*{grad}%
s^{\prime}-R^*\Vert_{1,w}.
	\end{multline*}
The deviation from the minimum norm $l_{2}$-minimizer $s^{\ast}$ is a
`median gradient flow' extracted from the
cyclic residual $R^{\ast}$, which moves the $l_{1}$-residual $\bar
{Y}-\operatorname*{grad}(s^*+\tilde{s})$ \textit{outside} the space of
divergence-free flows; here
\[
\tilde{s}\in\operatorname*{argmin}\nolimits_{s^\prime\in C^{0}}\lVert\operatorname*{grad}%
s^\prime-R^{\ast}\rVert_{1,w}.
\]

On the other hand, in the dual problem to \eqref{eq:l1}, we search for a
solution \textit{inside} the space of divergence-free flows. More precisely,
the dual form of the $l_{1}$-projection \eqref{eq:l1} searches within a space
of bounded divergence-free flows for a flow that is maximally correlated with
$\bar{Y}$. Before we state this theorem, we note that the inner product
defined in \eqref{eq:ip1} for skew-symmetric matrices representing edge
flows,
\[
\langle X,Y\rangle_{w}:=\sum\nolimits_{\{i,j\}\in E}w_{ij}X_{ij}Y_{ij},
\]
also defines an inner product over $\mathbb{R}^{n\times n}$ if the symmetric
weight matrix $W=[w_{ij}]$ has no zero entries, i.e.\ $w_{ij}>0$ for all
$i,j$. We will assume that this is the case in the following proposition.

\begin{proposition}
\label{thm:duall1} The $l_{1}$-projection problem \eqref{eq:l1} has the
following dual problem,
\begin{equation}%
\begin{array}
[c]{rcl}%
\max &  & \langle X,\bar{Y}\rangle_{w}\\
\mathrm{s.t.} &  & |X_{ij}|\leq1,\\
&  & \operatorname*{div}X=0,\\
&  & X=-X^{\top}.
\end{array}
\label{eq:l1dual}%
\end{equation}

\end{proposition}

\begin{proof}
This follows from standard duality theory for linear programming. See
\cite{Ye} for example.
\end{proof}

Proposition \ref{thm:duall1} shows that for $l_{1}$-projections, the dual
problem searches in the orthogonal complement of the primal domain. The primal
search space is the space of gradient flows $\operatorname*{im}%
(\operatorname*{grad})$ while the dual search space is the space of
divergence-free flows $\ker(\operatorname*{div})$. Recall that for $l_{2}%
$-projections, gradient flows correspond to the solutions while
divergence-free flows correspond to the residuals. So the solution-residual
split in the $l_{2}$-setting is in this sense analogous to the primal-dual
split in $l_{1}$-setting.

An optimal $l_{1}$-minimizer of \eqref{eq:l1} can only be decided up to a
constant from the complementary conditions,
\[
0<|X_{ij}|<1 \Rightarrow s_{j} - s_{i} = \bar{Y}_{ij}.
\]
The constraint $\sum_{i} s_{i} = 0$ may be imposed to remove this extra degree
of freedom.

\subsection{Conflict Identification: $l_{1}$-minimization for approximate
sparse cyclic rankings\label{sparse}}

In the discussion at the end of Section \ref{structure}, we mentioned that an
$l_{1}$-approximate sparse cyclic ranking for $R^{\ast}$ may be formulated as
the following $l_{1}$-minimization problem,
\begin{equation}%
\begin{array}
[c]{rcl}%
\min &  & \Vert X-R^{\ast}\Vert_{1}\\
\mathrm{s.t.} &  & X=\operatorname*{curl}^{\ast}\Phi,\\
&  & X=-X^{\top}.
\end{array}
\label{eq:sparsegen}%
\end{equation}
This is equivalent to
\[
\min_{\Phi\in C^{2}}\Vert\operatorname*{curl}\nolimits^{\ast}\Phi-R^{\ast
}\Vert_{1}:=\sum\nolimits_{\{i,j\}\in E}|(\operatorname*{curl}\nolimits^{\ast
}\Phi)_{ij}-R_{ij}^{\ast}|,
\]
which is in turn equivalent to
\[
\min_{\Phi\in C^{2}}\Vert\operatorname*{curl}\nolimits^{\ast}\Phi
-\operatorname*{proj}\nolimits_{\ker(\Delta_{1})}\bar{Y}\Vert_{1},
\]
where $\operatorname*{proj}_{\ker\Delta_{1}}\bar{Y}$ is the harmonic component
in $R^{\ast}$. The chief motivation for this minimization problem has been
explained at the end of Section \ref{structure} --- we would like to identify
the edges of conflicting pairs in a pairwise comparison graph so that we may
have the option of removing them to get a globally consistent ranking.

\begin{figure}[ptb]
\centering
\includegraphics[width=\textwidth]{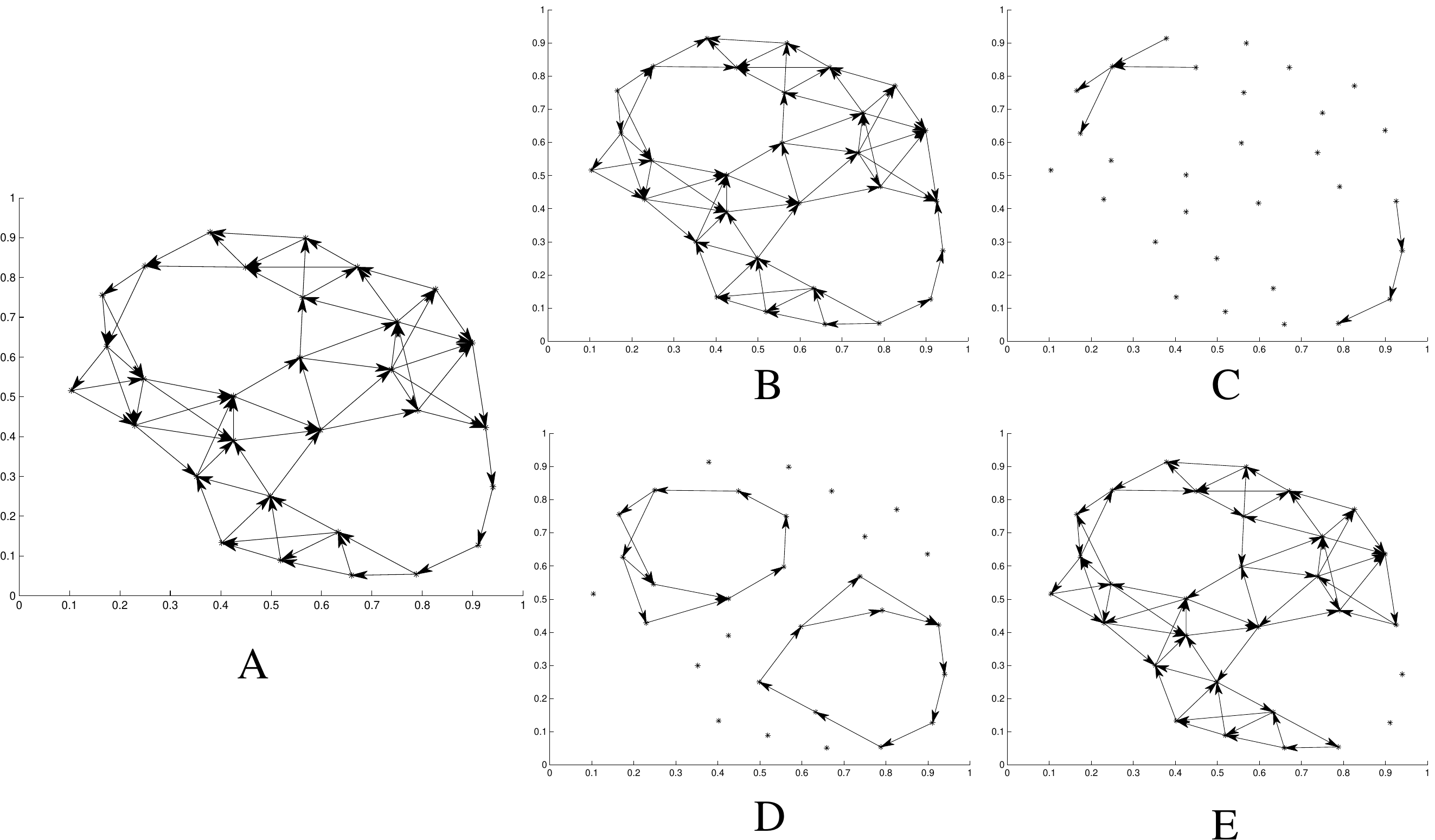} \caption{Comparisons
of the two $l_{1}$-optimizations, \eqref{eq:l1} and \eqref{eq:sparsegen}, with
the same harmonic ranking. For simplicity we set weights $w_{ij}=1$. The
arrows in the picture indicate the edge flow direction of pairwise rankings.
A. a harmonic ranking flow $h$; B. the $l_{1}$-projection on the gradient
flows by \eqref{eq:l1} (i.e.\ $\operatorname*{grad}s_{0}$ where $s_{0}%
=\operatorname*{argmin}_{s}\Vert\operatorname*{grad}s-h\Vert_{1}$); C. the
$l_{1}$-projection residual in \eqref{eq:l1} (i.e.\ $h-\operatorname*{grad}%
s_{0}$); D. the approximate sparse cycles by \eqref{eq:sparsegen}
(i.e.\ $h-\operatorname*{curl}^{\ast}\Phi_{0}$ where $\Phi_{0}%
=\operatorname*{argmin}_{\Phi}\Vert\operatorname*{curl}^{\ast}\Phi-h\Vert_{1}%
$); E. the $l_{1}$-projection on locally cyclic flows by \eqref{eq:sparsegen}
(i.e.\ $\operatorname*{curl}^{\ast}\Phi_{0}$). }%
\label{fig:harm_l1}%
\end{figure}

Both \eqref{eq:l1} and \eqref{eq:sparsegen} are $l_{1}$-norm minimizations
over some pairwise ranking flows. The main difference between them lies in
that the former model searches over $\operatorname*{im}(\operatorname*{grad}%
)$, the space of gradient flows, i.e.\ where $X=\operatorname*{grad}s$, while
the latter model searches over $\operatorname*{im}(\operatorname*{curl}^{\ast
})$, the space of curl flows, i.e.\ where $X=\operatorname*{curl}^{\ast}\Phi$.
The number of free parameters in $\operatorname*{grad}s$ is just $|V|=n$ but
the number of free parameters in $\operatorname*{curl}^{\ast}\Phi$ is
$|T(E)|$, which is typically of the order $O(n^{3})$. Therefore we expect to
be able to get a residual for \eqref{eq:sparsegen} that is much sparser than
the residual for \eqref{eq:l1} simply because we are searching over a much
larger space. As an illustration, Figure \ref{fig:harm_l1} shows the results
of these two optimization problems on the same data. 

The next proposition shows that the dual problem of \eqref{eq:sparsegen} also
maximizes correlation with the given pairwise ranking flow $R^{\ast}$ but over
bounded curl-free flows instead of bounded divergence-free flows as in \eqref{eq:l1dual}.

\begin{proposition}
\label{thm:dualsparse} Let the inner product be as defined in \eqref{eq:ip1},
i.e.
\[
\langle X,Y\rangle_{w}:=\sum\nolimits_{\{i,j\}\in E}w_{ij}X_{ij}Y_{ij}.
\]
The dual problem of the $l_{1}$-minimization \eqref{eq:sparsegen} is
\[%
\begin{array}
[c]{rcl}%
\max &  & \langle X,R^{\ast}\rangle_{w}\\
\mathrm{s.t.} &  & |X_{ij}|\leq w_{ij}^{-1},\\
&  & \operatorname*{curl}X=0,\\
&  & X=-X^{\top}.
\end{array}
\]

\end{proposition}

\begin{proof}
Similar to Proposition \ref{thm:duall1} with $\operatorname*{grad}$ replaced
by $\operatorname*{curl}^{\ast}$.
\end{proof}

As we can see, $\operatorname*{curl}$ in Proposition \ref{thm:dualsparse}
plays the role of $\operatorname*{div}$ in Proposition \ref{thm:duall1} in the
dual problem and $\operatorname*{curl}^{\ast}$ in Proposition
\ref{thm:dualsparse} plays the role of $\operatorname*{grad}$ in Proposition
\ref{thm:duall1} in the primal problem. There is a slight difference on the
upper bounds for $|X_{ij}|$, due to the fact that \eqref{eq:l1} uses a
weighted $l_{1}$-norm while \eqref{eq:sparsegen} uses an unweighted $l_{1}%
$-norm. In both propositions, the primal and dual search spaces are orthogonal
complements of each other as given by the Helmholtz decomposition theorem.
These two problems thus exhibit a kind of structural duality.

\section{Connections to Social Choice Theory\label{social}}

Social choice theory is almost undoubtedly the discipline most closely
associated with the study of ranking, having a long history dating back to
Condorcet's famous treatise in 1785 \cite{Condorcet1785} and a large body of
work that led to at least two Nobel prizes \cite{Arrow92,Sen98}.

The famous impossibility theorems of Arrow \cite{Arrow51} and Sen \cite{Sen70}
in social choice theory formalized the inherent difficulty of achieving a
global ranking of alternatives by aggregating over the voters. However it is
still possible to perform an approximate rank aggregation in reasonable,
systematic manners. Among the various proposed methods, the best known ones
are those by Condorcet \cite{Condorcet1785}, Borda \cite{Borda1781}, and
Kemeny \cite{Kemeny59}. In particular, the Kemeny approach is often regarded
as the best approximate rank aggregation method under some assumptions
\cite{YouLev78,Young88}. It is however NP-hard to compute and its sole
reliance on ordinal information may be unnatural in the context of score-based
cardinal data.

We have described earlier how the minimization of \eqref{eq:main2} over the
gradient flow model class
\[
\mathcal{M}_{G}= \{X \in C^{1} \mid X_{ij} = s_{j} - s_{i},\quad
s:V\to\mathbb{R}\}
\]
leads to a Hodge theoretic generalization of Borda count but the minimization
of \eqref{eq:main2} over the Kemeny model class
\[
\mathcal{M}_{K}= \{ X \in C^{1} \mid X_{ij} = \operatorname*{sign} (s_{j} -
s_{i}),\quad s: V\to\mathbb{R}\}
\]
leads to Kemeny optimization. In this section, we will discuss this connection
in greater detail.

The following are some desirable properties of ranking data that have been
widely studied, used, and assumed in social choice theory. A ranking problem
is called \textit{complete} if each voter in $\Lambda$ gives a total ordering
or permutation of all alternatives in $V$; this implies that $w^{\alpha}%
_{ij}>0$ for all $\alpha\in\Lambda$ and all distinct $i,j \in V$, in the
terminology of Section \ref{main}. It is \textit{balanced} if the pairwise
comparison graph $G=(V,E)$ is $k$-regular with equal weights $w_{ij} = c$ for
all $\{i,j\}\in E$. A complete and balanced ranking induces a complete graph
with equal weights on all edges. Moreover, it is \textit{binary} if every
pairwise comparison is allowed only two values, say, $\pm1$ without loss of
generality. So $Y^{\alpha}_{ij} = 1$ if voter $\alpha$ prefers alternative $j$
to alternative $i$, and $Y^{\alpha}_{ ij} = -1$ otherwise. Ties are disallowed
to keep the discussion simple.

Classical social choice theory often assumes complete, balanced, and binary
rankings. However, these are all unrealistic assumptions for modern data
coming from internet and e-commerce applications. Take the Netflix dataset for
illustration, a typical user $\alpha$ of Netflix would have rated at most a
very small fraction of the entire Netflix inventory. Indeed, as we have
mentioned in Section \ref{exmp:netflix}, the viewer-movie rating matrix has
$99\%$ missing values. Moreover, while blockbuster movies would receive a
disproportionately large number of ratings, since just about every viewer has
watched them, the more obscure or special interest movies would receive very
few ratings. In other words, the Netflix dataset is highly incomplete and
highly imbalanced. Therefore its pairwise comparison graph is expected to have
a sparse edge structure if we ignore pairs of movies where few comparisons
have been made\footnote{This will not be true if we do not perform such
thresholding. As we noted earlier, the Netflix pairwise comparison graph is
almost a complete graph missing only 0.22\% of its edges although the Netflix
dataset has 99\% of its values missing.}.

Lastly, as we have discussed in Section \ref{exmp:pairwise}, most modern
ranking datasets including the Netflix one are given in terms of ratings or
scores on the alternatives by the voters (e.g.\ one through five stars). While
it is possible to ignore the cardinal nature of the dataset and just use its
ordinal information to construct a binary pairwise ranking, we would be losing
valuable information --- for example, a 5-star versus 1-star comparison is
indistinguishable from a 3-star versus 2-star comparison when one only takes
the ordinal information into account.

Therefore, one is ill-advised to apply methods from classical social choice
theory to modern ranking data directly. We will see in the next section that
our Hodge theoretic extension of Borda count adapts to these new features in
modern datasets, i.e.\ incomplete, imbalanced, cardinal data, but still
restricts to the usual Borda count in social choice theory for data that is
complete, balanced, and ordinal/binary.

The reader may wonder why the impossibility theorems of social choice theory
do not invalidate our Hodge theoretic approach. One reason is given in the
previous paragraph, namely, we work under different assumptions: our ranking
data is incomplete, imbalanced, cardinal, and so these impossibility results
do not apply. In particular, these impossibility theorems are about
\textit{intransitivity}, i.e.\ whether one might have $i \succeq j \succeq k
\succeq i$, which is an ordinal condition; but our approach deals with
\textit{inconsistency}, i.e.\ whether one might have $X_{ij} + X_{jk} + X_{ki}
\ne0$, which is a cardinal condition. The second and more important reason is
that we do not merely seek a global ranking but also a locally cyclic ranking
and a harmonic ranking, with the latter two components accounting for the
cyclic inconsistencies in the ranking data. We acknowledge at the outset that
not all datasets can be reasonably assigned a global ranking but can sometimes
be cyclic in nature. So we instead seek to analyze ranking data by examining
its three constituting components: global, local, harmonic. The magnitude of
the cyclic (local + harmonic) component then quantifies the inconsistencies
that impede a global ranking. We do not always regard the cyclic component,
which measures the cardinal equivalent of the impossibilities in social choice
theory, as noise. In our framework, the data may be `explained' by a global
ranking only when the cyclic component is small; if that is not the case, then
the cyclic component is an integral part of the ranking data and one has no
reason to think that the global component would be any more informative than
the cyclic component.

\subsection{Kemeny Optimization and Borda Count}

The basic idea of Kemeny's rule \cite{Kemeny59,Kemeny-Snell73} is to minimize
the number of pairwise mismatches from a given ordering of the alternatives to
a voting profile, i.e.\ the collection of total orders on the alternatives by
each voter. The minimizers are called the \textit{Kemeny optima} and are often
regarded as the most reasonable candidates for a global ranking of the
alternatives. To be precise, we define the binary pairwise ranking associated
with a permutation $\sigma\in\mathfrak{S}_{n}$ (the permutation group on $n$
elements) to be $Y^{\sigma}_{ij}=\operatorname*{sign}(\sigma(i)-\sigma(j))$.
Given two total orders or permutations on the $n$ alternatives, $\sigma,
\tau\in\mathfrak{S}_{n}$, the \textit{Kemeny distance} (also known as
\textit{Kemeny-Snell} or \textit{Kendall $\tau$ distance}) is defined to be
\[
d_{K} (\sigma,\tau) := \frac{1}{2} \sum\nolimits_{i<j} |Y^{\sigma}_{ij} -
Y^{\tau}_{ij}| = \frac{1}{4} \sum\nolimits_{i,j} |Y^{\sigma}_{ij} - Y^{\tau
}_{ij}|,
\]
i.e.\ the number of pairwise mismatches between $\sigma$ and $\tau$. Given a
voting profile as a set of permutations on $V = \{ 1,\dots,n\}$ by $m$ voters,
$\{\tau_{i} \in\mathfrak{S}_{n} \mid i=1,\dots, m \}$, the following
combinatorial minimization problem
\begin{equation}
\label{eq:kemeny}\min_{\sigma\in\mathfrak{S}_{n} } \sum\nolimits_{i=1}^{m}
d_{K}(\sigma, \tau_{i})
\end{equation}
is called \textit{Kemeny optimization} and is known to be NP-hard
\cite{Dwork01} with respect to $n$ when $m \ge4$. For binary-valued rankings
with $Y^{\alpha}_{ij} \in\{\pm1\}$, the optimization problem
\begin{equation}
\label{eq:kopt}\min_{X\in\mathcal{M}_{K}} \sum\nolimits_{\alpha,i,j}
w^{\alpha}_{ij} (X_{ij} - Y^{\alpha}_{ij} )^{2} ,
\end{equation}
counts up to a constant the number of pairwise mismatches from a total order.
Hence for a complete, balanced, and binary-valued ranking problem, our
minimization problem \eqref{eq:main2} becomes Kemeny optimization if we
replace the subspace $\mathcal{M}_{G}$ by the discrete subset $\mathcal{M}%
_{K}$.

Another well-known method for rank aggregation is the \textit{Borda count}
\cite{Borda1781}, which assigns a voter's top $i$th alternative a
\textit{position-based score} of $n-i$; the global ranking on $V$ is then
derived from the sum of its scores over all voters. This is equivalent to
saying that the global ranking of the $i$th alternative is derived from the
score
\begin{equation}
\label{eq:borda}s_{B}(i) = -\sum\nolimits_{\alpha,k=1}^{m,n} Y^{\alpha}_{ik},
\end{equation}
i.e.\ the alternative that has the most pairwise comparisons in favor of it
from all voters will be ranked first, and so on. As we have found in
\eqref{eq:borda0}, the minimum norm solution of the $l_{2}$-projection onto
gradient flows is given by
\[
s^{*}(i) = - \frac{1}{n} \sum\nolimits_{k} \bar{Y}_{ik} = - c \sum
\nolimits_{\alpha,k=1}^{m,n} Y^{\alpha}_{ik},
\]
where $c$ is a positive constant. Hence for a complete, balanced, and binary
ranking problem, the Hodge theoretic approach yields the Borda count up (to a
positive multiplicative constant that has no effect on the ordering of
alternatives by scores).

\subsection{Comparative Studies\label{compare}}

The following theorem gives three equivalent characterizations of
\eqref{eq:kopt} when $Y^{\alpha}_{ij}\in\{\pm1\}$. Note that here we do not
assume that the data is complete and balanced.

\begin{theorem}
\label{thm7} Suppose that $Y^{\alpha}_{ij} \in\{\pm1\}$. The following
optimization problems are all equivalent:

\begin{enumerate}
[(i)]

\item \label{H} The weighted least squares problem,
\[
\min_{X\in\mathcal{M}_{K}} \sum\nolimits_{\alpha,i,j} w^{\alpha}_{ij} (X_{ij}
- Y^{\alpha}_{ij})^{2},
\]
where
\[
\mathcal{M}_{K} = \{ X \in\mathcal{A} \mid X_{ij} = \operatorname*{sign}(s_{j}
- s_{i}), \quad s:V\to\mathbb{R} \}.
\]

\item \label{I} The linear programming problem,
\begin{equation}
\label{eq:k1opt}\max_{X \in\mathcal{K}_{1}} \langle X, \bar{Y}\rangle= \max_{X
\in\mathcal{K}_{1}} \sum\nolimits_{\{i,j\}\in E} w_{ij} X_{ij} \bar{Y}_{ij},
\end{equation}
where $\mathcal{K}_{1}$ is the set
\[
\Bigl\{ \textstyle{\sum_{\sigma\in S_{n}}} \mu_{\sigma}P^{\sigma}
\Bigm| \textstyle{\sum_{\sigma}} \mu_{\sigma}=1,\quad\mu_{\sigma}\ge0,\quad
P^{\sigma}_{ij}=\operatorname*{sign}(\sigma(j)-\sigma(i)) \Bigr\}.
\]

\item \label{J} The weighted $l_{1}$-minimization problem,
\begin{equation}
\label{eq:k2opt}\min_{X\in\mathcal{K}_{2}} \lVert X - \bar{Y} \rVert_{1,w} =
\min_{X\in\mathcal{K}_{2}} \sum\nolimits_{\{i,j\} \in E} w_{ij} \lvert X_{ij}
- \bar{Y}_{ij} \rvert,
\end{equation}
where $\mathcal{K}_{2}$ is the set
\[
\{ X \in\mathcal{A} \mid(s_{j} - s_{i})X_{ij} \ge0\text{ for some } s:
V\to\mathbb{R} \text{ and } \{i,j\}\in E \}.
\]

\item \label{K} The minimum feedback arc set of the weighted directed graph
$G_{W\circ\bar{Y}}=(V, \vec{E}, W\circ\bar{Y})$, whose vertex set is $V$,
directed edge $(i,j)\in\vec{E} \subseteq V \times V$ iff $\bar{Y}_{ij}> 0$
with weight $w_{ij}\bar{Y}_{ij}$.
\end{enumerate}
\end{theorem}

\begin{proof}
Assuming (\ref{H}). Since $X_{ij} \in\{\pm1\}$, we obtain
\begin{align*}
\sum\nolimits_{\alpha,i,j} w^{\alpha}_{ij} (X_{ij} - Y^{\alpha}_{ij})^{2}  &
= \sum\nolimits_{\alpha,i,j} w^{\alpha}_{ij} \left[  X_{ij}^{2} - 2 X_{ij}
Y^{\alpha}_{ij} + (Y^{\alpha}_{ij})^{2} \right] \\
&  = c - 2\sum\nolimits_{i,j} X_{ij} \sum\nolimits_{\alpha} w^{\alpha}_{ij}
Y^{\alpha}_{ij}\\
&  = c - 2 \sum\nolimits_{i,j} w_{ij} X_{ij} \bar{Y}_{ij}%
\end{align*}
where $c$ is a constant that does not depend on $X$. So the problem becomes
\begin{equation}
\label{eq:LP}\max_{X\in\mathcal{M}_{K}} \sum\nolimits_{\{i,j\}\in E} w_{ij}
X_{ij} \bar{Y}_{ij}.
\end{equation}
Since $\mathcal{M}_{K}$ is a discrete set containing $n!$ points, a linear
programming problem over $\mathcal{M}_{K}$ is equivalent to searching over its
convex hull, i.e.\ $\mathcal{K}_{1}$, which gives (\ref{I}).

(\ref{K}) can also be derived from \eqref{eq:LP}. Consider a weighted directed
graph $G_{W\circ\bar{Y}}$ where an edge $(i, j) \in\vec{E}$ iff $\bar{Y}%
_{ij}>0$, and in which case has weight $|w_{ij}\bar{Y}_{ij}|$. \eqref{eq:LP}
is equivalent to finding a directed acyclic graph by reverting a set of edge
directions whose weight sum is minimized. This is exactly the minimum feedback
arc set problem.

Finally, we show that (\ref{J}) is also equivalent to the minimum feedback arc
set problem. For any $X\in\mathcal{K}_{2}$, the transitive region, there is an
associated weighted directed acyclic graph $G_{W\circ X}$ where an edge $(i,j)
\in\vec{E}$ iff $X_{ij}>0$, and in which case has weight $|w_{ij}X_{ij}|$.
Note that an optimizer of \eqref{eq:k2opt} has either $X^{*}_{ij}=-X^{*}_{ji}
= \bar{Y}_{ij}$ or $X^{*}_{ij}=-X^{*}_{ji} = 0$ on an edge $\{i,j\}\in E$,
which is equivalent to the problem of finding a directed acyclic graph by
deleting a set of edges from $G_{W\circ\bar{Y}}$ such that the sum of their
weights is minimized. Again, this is exactly the minimum feedback arc set
problem.

\end{proof}

The set $\mathcal{K}_{1}$ is the convex hull of the \textit{skew-symmetric
permutation matrices} $P^{\sigma}$ as defined in \cite{YouLev78}. The set
$\mathcal{K}_{2}$ is called the \textit{transitive pairwise region} by Saari
\cite{Saari00}, which comprises $n!$ cones corresponding to each of the $n!$
permutations on $V$.

It is known that the minimum feedback arc set problem in (\ref{K}) is NP-hard,
and therefore, so are the other three. Moreover, (\ref{J}) provides us with
some geometric insights when we view it alongside with \eqref{eq:main2}, the
$l_{2}$-projection onto gradient flows $\mathcal{M}_{G} =\{X \in\mathcal{A}
\mid X_{ij} = s_{j} - s_{i}, s: V\to\mathbb{R}\}$ which we have seen to be a
Hodge theoretic extension of Borda count. We will illustrate their differences
and similarities pictorially via the following example borrowed from Saari
\cite{Saari00}.

\begin{figure}[ptb]
\centering
\includegraphics[width=0.5\textwidth]{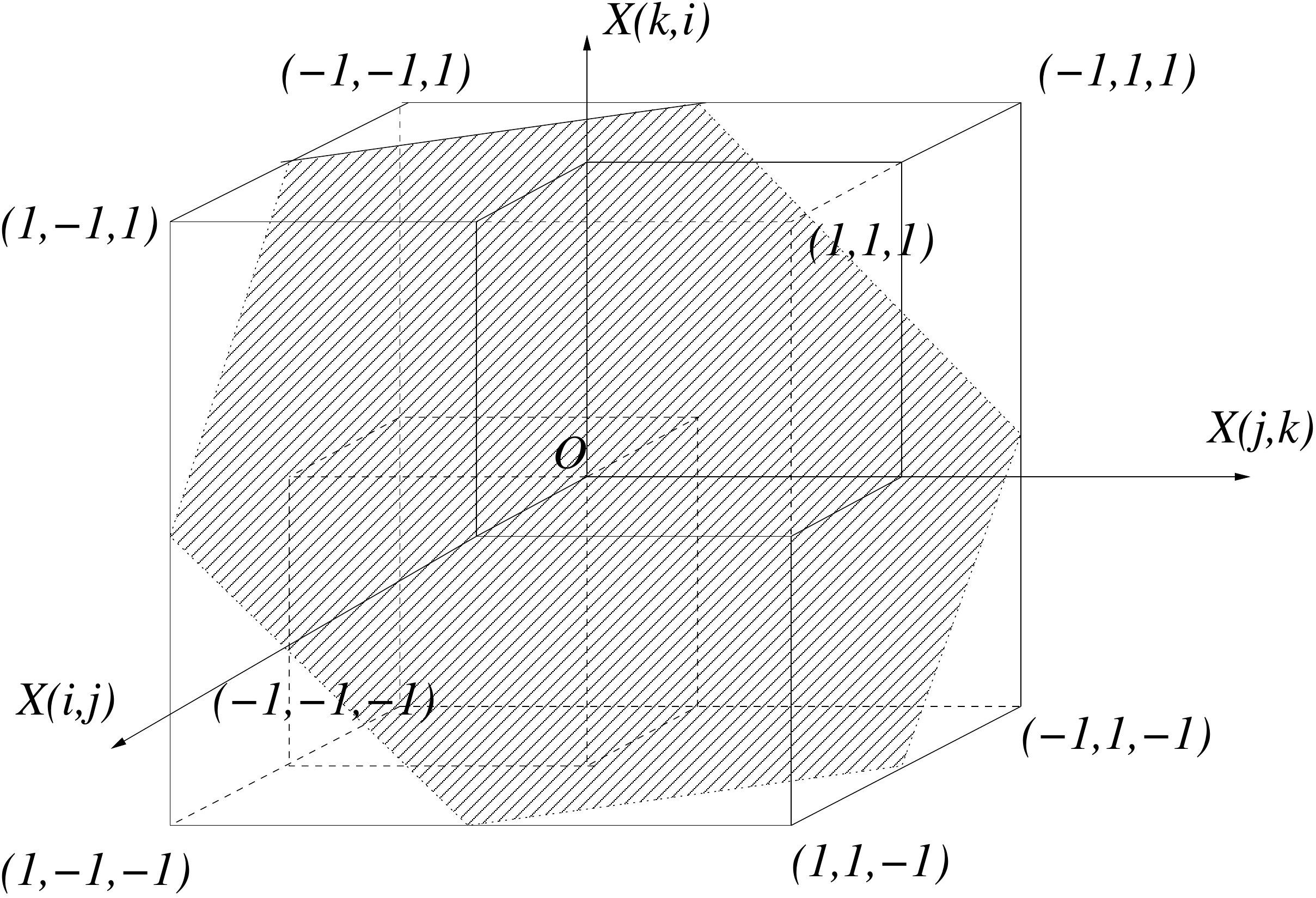}
\caption{The shaded region is the subspace $X_{ij}+X_{jk}+X_{ki}= 0$. The
transitive region consists of six orthants whose corresponding vertices belong
to $\{\pm1,\pm,1,\pm1\} - \{[1,1,1],[-1,-1,-1]\}$. The Borda count or
$\min_{X\in\mathcal{M}_{G}}I(X)$ is the $l_{2}$-projection onto the shaded
plane while the Kemeny optimization or $\min_{X\in\mathcal{M}_{K}}I(X)$ is the
$l_{1}$-projection onto the transitive region.}%
\label{fig:representCube}%
\end{figure}

Consider the simplest case of three-item comparison with $V=\{i,j,k\}$. For
simplicity, we will assume that $w_{ij}=w_{jk}=w_{ki}=1$ and $\bar{Y}%
_{ij},\bar{Y}_{jk},\bar{Y}_{ki} \in[-1,1]$. Figure \ref{fig:representCube}
shows the unit cube in $\mathbb{R}^{3}$. We will label the coordinates in
$\mathbb{R}^{3}$ as $[X_{ij},X_{jk},X_{ki}]$ (instead of the usual $[x,y,z]$).
The shaded plane corresponds to the set where $X_{ij} + X_{jk}+X_{ki}= 0$ in
the unit cube. Note that this set is equal to the model class $\mathcal{M}%
_{G}$ because of \eqref{eq:gradtrans}. On the other hand, the transitive
pairwise region $\mathcal{K}_{2}$ consists of the six orthants within the cube
with vertices $\{\pm1,\pm,1,\pm1\} - \{[1,1,1],[-1,-1,-1]\}$. We will write
\[
I(X): = \sum\nolimits_{\alpha,i,j} w^{\alpha}_{ij} (X_{ij} - Y^{\alpha}%
_{ij})^{2}.
\]
The Hodge theoretic optimization \eqref{eq:main2} is the $l_{2}$-projection
onto the plane $X_{ij}+X_{jk}+X_{ki}=0$, while by (\ref{J}), the Kemeny
optimization \eqref{eq:kopt} is the $l_{1}$-projection onto the aforementioned
six orthants representing the transitive pairwise region $\mathcal{K}_{2}$.

In the general setting of social choice theory, the following theorem from
\cite{Saari00} characterizes the order relations between the Kemeny
optimization and the Borda count.

\begin{theorem}
[Saari-Merlin 2000]The Kemeny winner (the most preferred) is always strictly
above the Kemeny loser (the least preferred) under the Borda count; similarly
the Borda winner is always strictly above the Borda loser under the Kemeny
rule. There is no other constraint in the sense that the two methods may
generate arbitrary different total orders except for those constraints.
\end{theorem}

The Kemeny rule has several desirable properties in social choice theory which
the Borda count lacks \cite{YouLev78}. The Kemeny rule satisfies the Condorcet
rule, in the sense that if an alternative in $V$ wins all pairwise comparisons
against other alternatives in $V$, then it must be the overall winner. A
Condorcet winner is any alternative $i$ such that $\sum_{j}
\operatorname*{sign}(\sum_{\alpha}Y^{\alpha}_{ij}) = n$. Note that the
Condorcet winner may not exist in general but Kemeny or Borda winners always
exist. However, if a Condorcet winner exists, then it must be the Kemeny
winner. On the other hand, Borda count can only ensure that the Condorcet
winner is ranked strictly above the Condorcet loser (least-preferred). Another
major advantage of the Kemeny rule is its consistency in global rankings under
the elimination of alternatives in $V$. The Borda count and many other
position-based rules fail to meet this condition. In fact, the Kemeny rule is
the unique rule that meets all three of following: (1) satisfies the Condorcet
rule, (2) consistency under elimination, and (3) a natural property called
neutral (that we will not discuss here). See \cite{YouLev78} for further details.

Despite the many important features that the Kemeny rule has, its high
computational cost (NP-hard) makes simpler rules like Borda count attractive
in practice, especially when there is large number of alternatives to be
ranked. Moreover, in cardinal rankings where it is desirable to preserve the
magnitude of score differences \cite{CorMohRas07} and not just the order
relation, using the Hodge theoretic variant of Borda count with model class
$\mathcal{M}_{G}$ becomes more relevant than Kemeny optimization with model
class $\mathcal{M}_{K}$.

\section{Experimental Studies\label{exp}}

We present three examples of Hodge theoretic ranking analysis of real data
with the hope that these preliminary results would illustrate some basic ideas
of our approach.

The first example is about movie ranking on a subset of Netflix data. We show
that (i) the use of pairwise ranking together with Hodge decomposition reduces
temporal drift bias, and (ii) the triangular curls provide a metric for
characterizing inconsistencies in the ranking data. The second example illustrates
the use of Hodge decomposition for finding a universal equivalent or price function
(i.e.\ global ranking) in a currency exchange market where triangular arbitrage-free
implies arbitrage-free (i.e.\ harmonic component is $0$).
The third example describes how the global ranking component in Hodge decomposition may be used to
approximate PageRank via reversible Markov chains.

\subsection{Movie Ranking on a Subset of Netflix Data\label{exp:netflix}}

The Netflix prize dataset\footnote{\url{http://www.netflixprize.com}} contains
about $17,000$ movies rated by $480,000$ customers over $74$ months from November
1998 to December 2005. Each customer rated $209$ movies on average and around
$99\%$ of the ratings are absent from the customer-product matrix. We do not
seek to address the Netflix prize problem of ratings prediction here. Instead
we take advantage of this rare publicly available dataset and use it to test
the rank aggregation capabilities of our method. We would like to aggregate
viewers' ratings into a global ranking on movies, and to measure the
reliability of such a global ranking. Note that such rank aggregation could be
\textit{personalized} if one first collects the ratings from viewers who share
similar tastes with an individual. This could then be used for rating
prediction if desired which is not pursued here.

\begin{figure}[ptb]
\centering
\includegraphics[width=\textwidth]{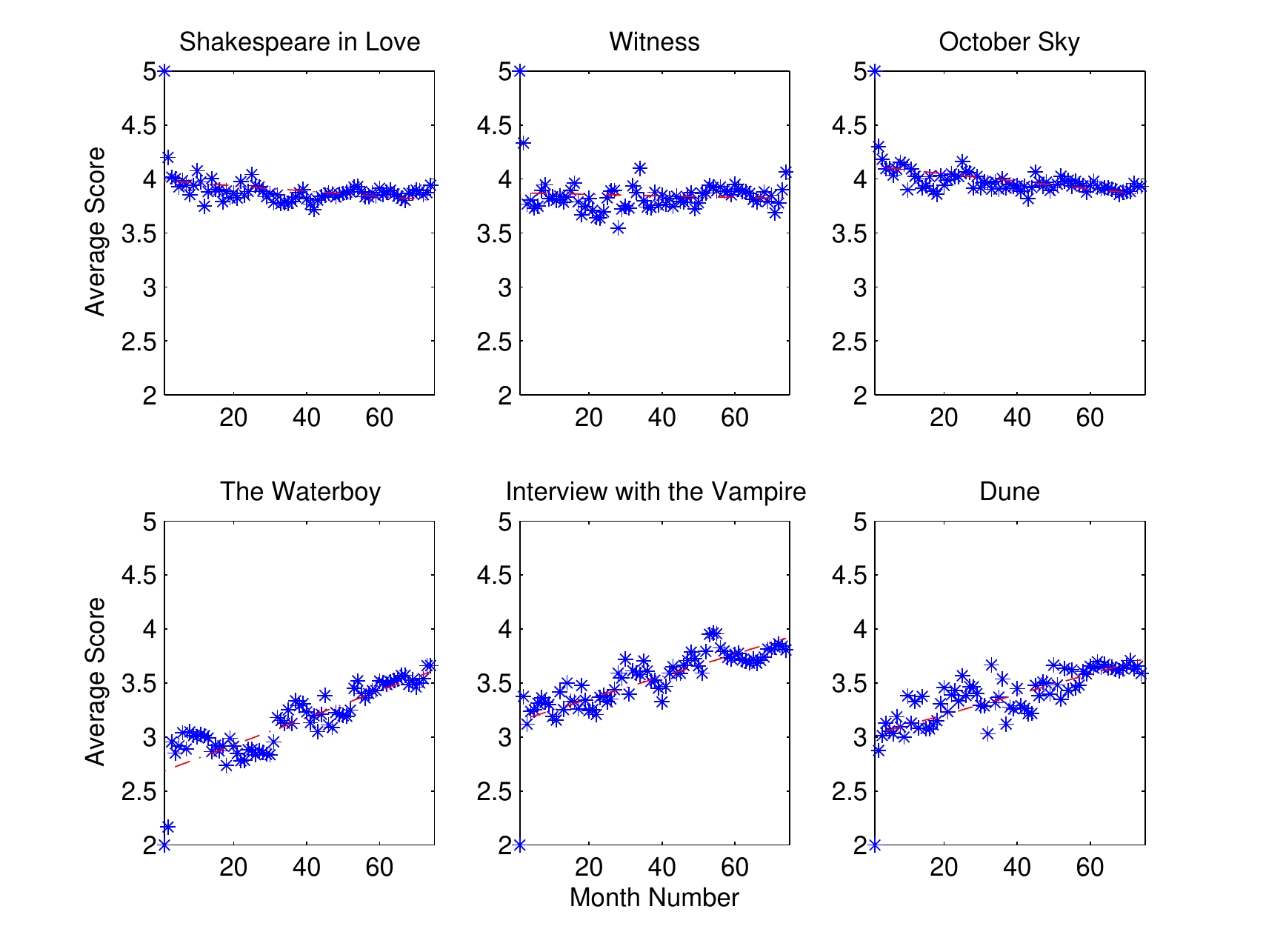}\caption{Average
scores of 6 selected movies over 74 months. The three movies in the top row
has a \textit{decreasing} trend in monthly average scores, while in a contrast
the other three movies in the bottom row exhibits an \textit{increasing}
trend.}%
\label{fig:movie6}%
\end{figure}

For reasons that we will soon explain, we restrict our selections to movies
that received ratings on all of the $74$ months. There are not many such
movies --- only $25$ in all. Several of these have monthly average scores that
show substantial upward or downward drifts. In Figure~\ref{fig:movie6}, we
show the temporal variations in scores of six of these (numerical indices in
the Netflix dataset are given in parentheses): \texttt{Dune} (17064), \texttt{Interview with the Vampire} (8079), \texttt{October Sky} (12473), \texttt{Shakespeare in Love} (17764), \texttt{The Waterboy} (14660), and \texttt{Witness} (15057). Such temporal variations make it
dubious to rank movies by simply taking average score over all users, as
ratings over different time periods may not be comparable under the same
scale. It is perhaps worth noting that understanding the temporal dynamics in
the Netflix dataset has been a key factor in the approach of Bell and Koren
\cite{KorBell07}. We will see below that the use of pairwise ranking and Hodge
decomposition provides an effective method to globally rank the movies and
detect any inherent inconsistency and that is furthermore robust under
temporal variations.

\begin{description}
\item[\textbf{Formation of pairwise ranking.}] Since pairwise rankings are
relative measures, we expect that they will reduce the effect of temporal
drift. We employ three of the statistics described in
Section~\ref{exmp:netflix} to form our pairwise rankings, using only ratings
by the same customer in the same month. We compute the \textit{arithmetic mean
of score differences},
\[
\bar{Y}_{ij}=\frac{\sum_{\alpha}(a_{\alpha j}-a_{\alpha i})}{\#\{\alpha\mid
a_{\alpha i},a_{\alpha j}\text{ exist in the same month}\}};
\]
the \textit{geometric mean of score ratios},
\[
\bar{Y}_{ij}=\frac{\sum_{\alpha}(\log a_{\alpha j}-\log a_{\alpha i}%
)}{\#\{\alpha\mid a_{\alpha i},a_{\alpha j}\text{ exist in the same month}\}};
\]
and \textit{binary comparisons},
\begin{multline*}
\bar{Y}_{ij}=\Pr\{\alpha\mid a_{\alpha j}>a_{\alpha i}\}-\Pr\{\alpha\mid
a_{\alpha j}<a_{\alpha i}\},\\
\text{where }\alpha\text{ is such that }a_{\alpha i},a_{\alpha j}\text{ exist
in the same month}.
\end{multline*}
Since there is nothing to suggest that a logarithmic scale is relevant, the
logarithmic odds ratio gives rather poor result as expected and we omitted it.
For comparison, we compute the mean score of each movie over all customers,
ignoring the temporal information. A reference score is collected
independently from MRQE (Movie Review Query
Engine)\footnote{\url{http://www.mrqe.com}}, the largest online directory of
movie reviews on the internet.

\item[\textbf{Global ranking by Hodge decomposition.}] We then solve the
regression problem in \eqref{eq:main2} to obtain a projection of pairwise
ranking flows onto gradient flows, given by Theorem~\ref{thm:l2}(\ref{A}).
Note that in this example, the pairwise ranking graph is complete with $n=6$
nodes. Table~\ref{tab:nflix} collects the comparisons between different global
rankings. The reference order of movies is again via the MRQE scores.

\item[\textbf{Inconsistencies and curls.}] Since the pairwise ranking graph is
complete, its clique complex is a simplex with $n=6$ vertices and so the
harmonic term in the Hodge decomposition is always zero. Hence the residual in
Theorem~\ref{thm:l2} is just the curl projection, i.e.\ $R^{\ast
}=\operatorname*{proj}_{\operatorname*{im}(\operatorname*{curl}^{\ast})}%
\bar{Y}$. We will define two indices of inconsistency to evaluate the results.
The first, called \textit{cyclicity ratio}, is a measure of global
inconsistency given by
\[
C_{p}=\frac{\Vert R^{\ast}\Vert_{2,w}^{2}}{\Vert\bar{Y}\Vert_{2,w}^{2}};
\]
while the second, called \textit{relative curl}, quantifies the local
inconsistency, and is given by the following function of edges and triangles,
\[
c_{r}(e_{ij},t_{ijk})=\frac{(\operatorname{curl}\bar{Y})(i,j,k)}%
{3(\operatorname{grad}s^{\ast})(i,j)}=\frac{\bar{Y}_{ij}+\bar{Y}_{jk}+\bar
{Y}_{ki}}{3(s_{j}^{\ast}-s_{i}^{\ast})}.
\]
Note that on every triangle $t_{ijk}$ the curl $\bar{Y}_{ij}+\bar{Y}_{jk}%
+\bar{Y}_{ki}$ measures the total sum of cyclic flow, therefore $c_{r}$
measures the magnitude of its induced edge flow relative to the gradient edge
flow of the global ranking $s^{\ast}$. If $c_{r}$ has absolute value larger
than $1$, then the average cyclic flow has an effect larger than the global
ranking $s^{\ast}$, which indicates that the global ranking $s^{\ast}$ might
be inconsistent on the pair of items.
\end{description}

Table~\ref{tab:nflix} shows that in terms of cyclicity ratio, the best global
ranking is obtained from Hodge decomposition of pairwise rankings from binary
comparisons, which has the smallest cyclicity ratio, $0.30$. This global
ranking is quite different from merely taking mean scores and a better
predictor of MRQE.

A closer analysis of relative curls allows us to identify the dubious scores.
We will see that the placement of \texttt{Witness} and \texttt{October Sky} according to the global ranking
contains significant inconsistency and should not be trusted. This inconsistency is largely due
to the curls in the triangles
\begin{align*}
t_1 &= (\text{\texttt{Witness}}, \text{\texttt{October Sky}}, \text{\texttt{The Waterboy}}),\\
t_2 &= (\text{\texttt{Witness}}, \text{\texttt{October Sky}}, \text{\texttt{Interview with the Vampire}}).
\end{align*}
In fact, there are only two relative curls whose magnitudes exceed $1$; both occurred on triangles that contain the edge $e = (\text{\texttt{Witness}}, \text{\texttt{October Sky}})$: The relative curl of $t_1$ with respect to $e$ is $3.6039$ while that of $t_2$ with respect to $e$ is $4.1338$. As we can see from Table~\ref{tab:nflix}, the inconsistency (large curl) manifests itself as \textit{instability} in the placement of \texttt{Witness} and \texttt{October Sky} --- the results vary across different rank aggregation methods with no possibility of consensus. This illustrates the use of curl as a certificate of validity for global ranking.

\begin{table}[ptbh]
\begin{center}
\resizebox{\textwidth}{!}{
\begin{tabular}
[c]{@{}cccccc@{}}%
\toprule & \multicolumn{5}{c}{Global ranking (Score)}\\
\cmidrule(l){2-6} Movie & MRQE & Mean & Hodge-Difference & Hodge-Ratio &
Hodge-Binary\\
\midrule \texttt{Shakespeare in Love} & 1 (85) & 2 ($3.87$) & 1 ($0.247$) & 2
($0.0781$) & 1 ($0.138$)\\
\texttt{Witness} & 2 ($77$) & 3 ($3.86$) & 2 ($0.217$) & 1 ($0.0883$) & 3 ($0.107$)\\
\texttt{October Sky} & 3 ($76$) & 1 ($3.93$) & 3 ($0.213$) & 3 ($0.0775$) & 2
($0.111$)\\
\texttt{The Waterboy} & 4 ($66$) & 6 ($3.38$) & 6 ($-0.464$) & 6 ($-0.1624$) & 6
($-0.252$)\\
\texttt{Interview with the Vampire} & 5 ($65$) & 4 ($3.71$) & 4 ($-0.031$) & 4
($-0.0121$) & 4 ($-.012$)\\
\texttt{Dune} & 6 ($44$) & 5 ($3.49$) & 5 ($-0.183$) & 5 ($-0.0693$) & 5 ($-0.092$%
)\\\hline
Cyclicity ratio & -- & -- & $0.77$ & $1.15$ & $0.30$\\
\bottomrule\\
\end{tabular}}
\end{center}
\caption{Global ranking of selected six movies via different methods: MRQE,
mean score over customers, Hodge decomposition with algorithmic mean score
difference, Hodge decomposition with geometric mean score ratio, and Hodge
decomposition with binary comparisons. It can be seen that the Hodge
decomposition with binary comparisons has the smallest inconsistency in terms
of the cyclicity ratio.}%
\label{tab:nflix}%
\end{table}

\subsection{Currency Exchange Market\label{exp:currency}}

This example illustrates a globally consistent pairwise ranking on a complete
graph using currency exchange data taken from
Yahoo!~Finance\footnote{\url{http://finance.yahoo.com/currency-converter}}.
Consider a currency exchange market with $V$ representing a collection of
seven currencies, USD, JPY, EUR, CAD, GBP, AUD, and CHF. In this case,
$G=(V,E)$ is a complete graph since every two currencies in $V$ are
exchangeable. Table~\ref{tab:currency} shows the exchange rates. By
logarithmic transform the exchange rates can be converted into pairwise
rankings as in Example~\ref{exmp:currency}. The global ranking is the solution
in \eqref{eq:glbrank} (where $\delta_{0}^{\ast}=\delta_{0}^{\top}$) defines an
\textit{universal equivalent} which measures the `value'\ of each currency. As
the reader can easily check, the logarithmic transform of the data in
Table~\ref{tab:currency} is curl-free (up to machine precision), which in this
context means triangular arbitrage-free. In other words, there is no way one
could profit from a cyclic exchange of any three currencies in $V$. Since $G$
is a complete graph, the data has no harmonic components; so Hodge
decomposition tells us that local consistency must imply global consistency,
which in this context means arbitrage-free. In other words, there is no way
one could profit from a cyclic exchange of any number of currencies in $V$ either.

\begin{table}[ptb]
\begin{center}
\resizebox{\textwidth}{!}{
\begin{tabular}{@{} cccccccc @{}}
	\toprule
\multicolumn{8}{c}{Currency exchange rate table} \\
\cmidrule(l){2-8}
\textit{} & USD & JPY & EUR & CAD & GBP & AUD & CHF
\\ \midrule
1 USD $=$  & 1.0000 & 114.6700  &  0.6869  &  0.9187  &  0.4790  &  1.0768  &  1.1439 \\
1 JPY $=$ &  0.0087   & 1.0000  &  0.0060  &  0.0080  &  0.0042  &  0.0094 &   0.0100 \\
1 EUR $=$  &  1.4558 &  166.9365   & 1.0000   & 1.3374  &  0.6974  &   1.5676   & 1.6653 \\
1 CAD $=$ & 1.0885  & 124.8177  &  0.7477  &  1.0000  &  0.5214  &  1.1721  &  1.2451 \\
1 GBP $=$ & 2.0875 & 239.3791  &  1.4340  &  1.9178  &  1.0000  &  2.2478  &  2.3879 \\
1 AUD $=$ & 0.9287 & 106.4940  &  0.6379  &  0.8532  &  0.4449  &  1.0000  &  1.0623 \\
1 CHF $=$ & 0.8742 & 100.2448  &  0.6005  &  0.8031  &  0.4188  &  0.9413  &  1.0000 \\  \midrule
Universal equivalent &  1.7097   &  0.0149  &  2.4890  &  1.8610  &  3.5691  &  1.5878  &  1.4946 \\
\bottomrule\\
\end{tabular}}
\end{center}
\caption{The last line is given by $\exp(-x^{\ast})$ where $x^{\ast}$ is the
solution to \eqref{eq:glbrank}. The data was taken from the Currency Converter
Yahoo!~Finance on November 6, 2007.}%
\label{tab:currency}%
\end{table}

\subsection{Comparisons with PageRank and HITS\label{exp:pagerank}}

We apply Hodge theoretic ranking to the problem of web ranking, which we
assumed here to mean any static linked objects, not necessarily the World Wide
Web. As we shall see Hodge decomposition provides an alternative to PageRank
\cite{PageRank} and HITS \cite{HITS}. In particular, it gives a new way to
approximate PageRank and enables us to study the inconsistency or cyclicity in
PageRank models.

Consider a link matrix $L$ where $L_{ij}$ is the number of links from site $i$
to $j$. There are two well-known spectral approach to computing the global
rankings of websites from $L$, HITS and PageRank. HITS computes the singular
value decomposition $L=U\Sigma V^{\top}$, where the primary left-singular
vector $u_{1}$ gives the \textit{hub ranking} and the primary right-singular
vector $v_{1}$ gives the \textit{authority ranking} (both $u_{1}$ and $v_{1}$
are nonnegative real-valued by the Perron-Frobenius theorem). PageRank
constructs from $L$ a Markov chain on the sites given by%
\[
P_{ij}=\alpha\frac{L_{ij}}{\sum_{j}L_{ij}}+(1-\alpha)\frac{1}{n},
\]
where $n$ is the number of sites and $\alpha=0.85$ trades-off between
Markovian link jumps and random surfing.

It is clear that we may define an edge flow via%
\begin{equation}
Y_{ij}=\log\frac{P_{ij}}{P_{ij}}. \label{eq:pageflow}%
\end{equation}
However what property does such a flow capture in PageRank? To answer this
question we will need to recall the notion of a reversible Markov chain: An
irreducible Markov chain with transition matrix $P$ and stationary
distribution $\pi$ is \textit{reversible} if
\[
\pi_{i}P_{ij}=\pi_{j}P_{ji}.
\]
Therefore a reversible Markov chain $P$ has a pairwise ranking flow induced
from a global ranking,
\[
Y_{ij}=\log\frac{P_{ij}}{P_{ij}}=\log\pi_{j}-\log\pi_{i},
\]
where $\log\pi$ gives the global ranking. As we mentioned in
Section~\ref{sec:Pairwise}, $\log\pi$ may be viewed as defining a
\textit{negative potential} on webpages if we regard ranking as being directed
from a higher potential site to a lower potential site. This leads to the
following interpretation.

Let $P^{\ast}$ be the best reversible approximate of the PageRank Markov chain
$P$, in the sense that%
\[
P^{\ast}=\operatorname*{argmin}\nolimits_{\tilde{P}\text{ reversible}%
}\left\Vert \log\frac{\tilde{P}_{ij}}{\tilde{P}_{ji}}-\log\frac{P_{ij}}%
{P_{ji}}\right\Vert _{2}.
\]
Then the stationary distribution of $P^{\ast}$, denoted by $\pi^{\ast}$, is a
Gibbs-Boltzmann distribution on webpages with potential $-s^{\ast}$, i.e.
\[
\pi_{i}^{\ast}=\frac{e^{s_{i}^{\ast}}}{\sum_{k}e^{s_{k}^{\ast}}}.
\]
where $s^{\ast}$ is given by the Hodge projection of $Y$ onto the space of
gradient flows. Hence the Hodge decomposition of edge flow in
\eqref{eq:pageflow} gives the stationary distribution of a best reversible
approximate of the PageRank Markov chain.

We may further compute the Hodge decomposition of iterated flows,
\[
Y_{ij}^{k}=\log\frac{P_{ij}^{k}}{P_{ij}^{k}}.
\]
Clearly when $k\rightarrow\infty$, the global ranking given by Hodge
decomposition converges to that given by PageRank. The benefit of the Hodge
theoretic approach lies in that (i) it provides a way to approximate the
PageRank stationary distribution; and (ii) it enables us to study the
inconsistency or cyclicity in PageRank Markov model. The cost of computing the
global ranking by Hodge decomposition in Theorem~\ref{thm:l2}(\ref{A}) only
involves a least squares problem of the graph Laplacian, which is less
expensive than eigenvector computations in PageRank. For the benefit of
readers unfamiliar with numerical linear algebra, it might be worth pointing
out that even the most basic algorithms for linear least squares problems
guarantee global convergence in a finite number of steps whereas there are (a)
no algorithms for eigenvalue problems that would terminate in a finite number
of steps as soon as the matrix dimension exceeds $4$; and (b) no algorithms
with guaranteed global convergence for arbitrary input matrices.

To illustrate this discussion, we use the UK Universities Web Link Structure
dataset\footnote{This is available from
\url{http://cybermetrics.wlv.ac.uk/database/stats/data}. We used counts at the
directory level.}. The dataset contains the number of web links between $111$
UK universities in 2002. Independent of this link structure is a research
score for each university, RAE 2001, performed during the 5-yearly Research
Assessment Exercise\footnote{http://www.rae.ac.uk}. The RAE scores are widely
used in UK for measuring the quality of research in universities. We used
$107$ universities by eliminating four that are missing either RAE score,
in-link, or out-link. The data has also been used by \cite{XuOwen09} recently
but for a different purpose. Table \ref{tab:page} summarizes the comparisons
among nine global rankings: RAE 2001, in-degree, out-degree, HITS authority,
HITS hub, PageRank, Hodge rank with $k=1,2$, and $4$, respectively. We then
use Kendall $\tau$-distance \cite{KenGib90} to count the number of pairwise mismatches between
global rankings, normalized by the total number of pairwise comparisons.

\begin{table}[ptb]
\begin{center}
\resizebox{0.95\textwidth}{!}{
\begin{tabular}{@{} ccccccccccc @{}}
	\toprule
\multicolumn{10}{c}{Kendall $\tau$-distance} \\
\cmidrule(l){2-10}
\textit{} & RAE'01 & in-degree & out-degree & HITS authority & HITS hub & PageRank  &  Hodge ($k=1$)  & Hodge ($k=2$) &  Hodge ($k=4$)
\\ \midrule
RAE'01 &        0  &  0.0994 &   0.1166  &  0.0961 &   0.1115  &  0.0969  &  0.1358  &  0.0975  &  0.0971 \\
in-degree & 0.0994& 0 & 0.0652& 0.0142& 0.0627& 0.0068& 0.0711& 0.0074& 0.0065 \\
out-degree & 0.1166& 0.0652& 0 & 0.0672& 0.0148& 0.0647& 0.1183& 0.0639& 0.0647 \\
HITS authority & 0.0961& 0.0142& 0.0672& 0 & 0.0627& 0.0119& 0.0736& 0.0133& 0.0120 \\
HITS hub & 0.1115& 0.0627& 0.0148& 0.0627& 0 & 0.0615& 0.1121& 0.0607& 0.0615 \\
PageRank & 0.0969& 0.0068& 0.0647& 0.0119& 0.0615& 0 & 0.0710& 0.0029& 0.0005 \\
Hodge ($k=1$) & 0.1358& 0.0711& 0.1183& 0.0736& 0.1121& 0.0710& 0 & 0.0692& 0.0709 \\
Hodge ($k=2$) & 0.0975& 0.0074& 0.0639& 0.0133& 0.0607& 0.0029& 0.0692& 0 & 0.0025 \\
Hodge ($k=3$) & 0.0971& 0.0065& 0.0647& 0.0120& 0.0615& 0.0005& 0.0709& 0.0025& 0 \\
\bottomrule\\
\end{tabular}}
\end{center}
\caption{Kendall $\tau$-distance between different global rankings. Note that
HITS authority gives the nearest global ranking to the research score RAE'01,
while Hodge decompositions for $k\geq2$ give closer results to PageRank which
is the second closest to the RAE'01. }%
\label{tab:page}%
\end{table}

\section{Summary and Conclusion\label{conclude}}

We introduced combinatorial Hodge theory to statistical ranking methods based
on minimizing pairwise ranking errors over a model space. In particular, we
proposed a Hodge theoretic approach towards determining the global, local, and
harmonic ranking components of a dataset of voters' scores on alternatives.
The global ranking is learned via an $l_{2}$-projection of a pairwise ranking
edge flow onto the space of gradient flows. We saw that among other
connections to classical social choice theory, the score recovered from this
global ranking is a generalization of the well-known Borda count to ranking
data that is cardinal, imbalanced, and incomplete. The residual left is the
$l_{2}$-projection onto the space of divergence-free flows. A subsequent
$l_{2}$-projection of this divergence-free residual onto the space of
curl-free flows then yields a harmonic flow. This decomposition of pairwise
ranking data into a global ranking component, a locally cyclic ranking
component, and a harmonic ranking component, is called the Helmholtz decomposition.

Consistency of the ranking data is governed to a large extent by the structure
of its pairwise comparison graph; this is in turn revealed in the Helmholtz
decomposition associated with the graph Helmholtzian, the combinatorial
Laplacian of the $3$-clique complex. The sparsity structure of a pairwise
comparison graph imposes certain constraints on the topology and geometry of
its clique complex, which in turn decides the properties of our statistical
ranking algorithms.

In addition one may use an $l_{1}$-approximate sparse cyclic rankings to
identify conflicts among voters. The $l_{1}$-minimization problem for this has
a dual given by correlation maximization over bounded curl-free flows. On the
other hand, the $l_{1}$-projection on the gradient flows, which we view as a
robust variant of the $l_{2}$-version, has a dual given by correlation
maximization over bounded cyclic flows.

Our results suggest that combinatorial Hodge theory could be a promising tool
for the statistical analysis of ranking, especially for datasets with
cardinal, incomplete, and imbalanced information.

\end{document}